\documentclass[11pt,a4paper]{article}
\usepackage[hyperref]{acl2021}
\usepackage{times}
\usepackage{ulem}
\usepackage{latexsym}

\usepackage{microtype}
\usepackage{multirow}

\aclfinalcopy 
\def\aclpaperid{2064} 

\usepackage{quoting}
\usepackage{xcolor}
\usepackage{microtype}
\usepackage{xspace}
\usepackage{graphicx}
\usepackage{subfig}
\usepackage{booktabs}
\usepackage{amsmath}
\usepackage{tikz}
\usepackage{todonotes}
\usepackage{enumitem}
\usetikzlibrary{plotmarks}

\newcommand\marksymbol[2]{\tikz[#2,scale=1.8]\pgfuseplotmark{#1};}

\newcommand{\Note}[2]{\todo[color=#1,size=\small, inline=true]{#2}} 
\newcommand{\SideNote}[2]{\todo[color=#1,size=\small]{#2}} %
\newcommand{\violet}[1]{\SideNote{purple!40}{#1 --violet}}
\newcommand{\violetil}[1]{\Note{purple!40}{#1 --violet}}
\newcommand{\mkclean}{
    \renewcommand{\violet}[1]{}
	\renewcommand{\violetil}[1]{}
}
\mkclean

\definecolor{emerald}{rgb}{0.31, 0.47, 0.26}
\definecolor{lightblue}{RGB}{53, 122, 191}
\definecolor{lightorange}{RGB}{226, 134, 49}

\newenvironment{myquote}[1]%
  {\list{}{\leftmargin=#1\rightmargin=#1}\item[]}%
{\endlist}

\title{Men Are Elected, Women Are Married: Events Gender Bias on Wikipedia}

\author{Jiao Sun$^{1,2}$ {\normalfont{and}} Nanyun Peng$^{1,2,3}$ \\
{$^1$Computer Science Department, University of Southern California}\\
{$^2$Information Sciences Institute, University of Southern California}\\
{$^3$Computer Science Department, University of California, Los Angeles}\\
\tt{\href{mailto:jiaosun@usc.edu}{jiaosun@usc.edu}},  \tt{\href{mailto:violetpeng@ucla.cs.edu}{violetpeng@ucla.cs.edu}}
}

\date{}

\begin{document}
\maketitle

\begin{abstract}

Human activities can be seen as sequences of events, which are crucial to understanding societies. 
Disproportional event distribution for different demographic groups 
can manifest and amplify social stereotypes, and potentially jeopardize the ability of members in some groups to pursue certain goals.
In this paper, we present the first event-centric study of gender biases in a Wikipedia corpus. To facilitate the study, we curate a corpus of career and personal life descriptions with demographic information consisting of 7,854 fragments from 10,412 celebrities. Then we detect events with a state-of-the-art event detection model, calibrate the results using strategically generated templates, and extract events that have asymmetric associations with genders.
Our study discovers that Wikipedia pages tend to intermingle personal life events with professional events for females but not for males, which calls for the awareness of the Wikipedia community to formalize guidelines and train the editors to mind the implicit biases that contributors carry. 
Our work also lays the foundation for future works on quantifying and discovering event biases at the corpus level.

\end{abstract}

\section{Introduction}


Researchers have been using NLP tools to analyze corpora for various tasks on online platforms. For example, ~\citet{Pei2020QuantifyingII} found that female-female interactions are more intimate than male-male interactions on Twitter and Reddit. 
Different from social media, open collaboration communities such as Wikipedia have slowly won the trust of public~\cite{Young2016ItsNW}. Wikipedia has been trusted by many, including professionals in work tasks such as scientific journals~\cite{Kousha2017AreWC} and public officials in powerful positions of authority such as court briefs~\cite{Gerken2010HowCU}. Implicit biases in such knowledge sources could have a significant impact on audiences' perception of different groups, thus propagating and even amplifying societal biases. Therefore, analyzing potential biases in Wikipedia is imperative.


\begin{table}[t]
\small
\centering
\begin{tabular}{@{ }p{1cm}@{ } | @{ }p{6.1cm}@{ }}
\toprule \textbf{Name}& \ \textbf{Wikipedia Description} \\ \midrule
Loretta Young ({\color{lightorange}{F}}) &   \ \begin{tabular}[c]{@{}p{6.1cm}@{}} \textbf{{\color{lightorange}Career}}: In 1930, when she was 17, she \colorbox{yellow!50}{eloped} with 26-year-old actor \underline{Grant Withers}; they were \colorbox{yellow!50}{married} in Yuma, Arizona. The \colorbox{yellow!50}{marriage} was \colorbox{yellow!50}{annulled} the next year, just as their second movie together (ironically entitled Too Young to Marry) was \colorbox{yellow!50}{released}. \end{tabular} \\
\midrule
Grant Withers ({\color{lightblue}M}) &   \ \begin{tabular}[c]{@{}p{6.1cm}@{}} 
\textbf{{\color{lightblue}Personal Life}}: In 1930, at 26, he \colorbox{yellow!50}{eloped} to Yuma, Arizona with 17-year-old actress Loretta Young. The \colorbox{yellow!50}{marriage} ended in \colorbox{yellow!50}{annulment} in 1931 just as their second movie together, titled Too Young to Marry, was \colorbox{yellow!50}{released}. \end{tabular} \\
\bottomrule
\end{tabular}
\caption{The marriage events are under the \emph{Career} section  for the female on Wikipedia. However, the same marriage is in the \emph{Personal Life} section for the male. \colorbox{yellow!50}{yellow} background highlights events in the passage.}
\label{table:wiki}
\end{table}

In particular, studying events in Wikipedia is important. 
An event is a specific occurrence under a certain time and location that involves participants~\cite{ACE}; human activities are essentially sequences of events. Therefore, the distribution and perception of events shape the understanding of society.
~\citet{Rashkin2018Event2MindCI} discovered implicit gender biases\violet{can you be more specific? What type of gender bias? } in film scripts using events as a lens. For example, they found that events with female agents are intended to be helpful to other people, while events with male agents are motivated by achievements.  However, they focused on the intentions and reactions of events rather than events themselves.

In this work, we propose to use events as a lens to study gender biases and demonstrate that events are more efficient for understanding biases in corpora than raw texts. 
We define \emph{gender bias} as the asymmetric association of events with females and males,\footnote{In our analysis, we limit to binary gender classes, which, while unrepresentative of the real-world diversity, allows us to focus on more depth in analysis.} which may lead to gender stereotypes. For example, females are more associated with domestic activities than males in many cultures~\cite{Leopold2018GenderDI,Jolly2014GenderDI}. 


To facilitate the study, we collect a corpus that contains demographic information, personal life description, and career description from Wikipedia.\footnote{\tiny\url{https://github.com/PlusLabNLP/ee-wiki-bias}}\violet{please put this under pluslab} 
We first detect events in the collected corpus using a state-of-the-art event extraction model~\cite{Han2019JointEA}.
Then, we 
extract gender-distinct events with a higher chance to occur for one group than the other. Next, we propose a calibration technique to offset the potential confounding of gender biases in the event extraction model, enabling us to focus on the gender biases at the corpus level. Our contributions are three-fold:

\begin{itemize}
    \item We contribute a corpus of 7,854 fragments from 10,412 celebrities across 8 occupations including their demographic information and Wikipedia \emph{Career} and \emph{Personal Life} sections.
    \item We propose using events as a lens to study gender biases at the corpus level\violet{add a brief description of what biases we found in the corpus.}, discover a mixture of personal life and professional life for females but not for males, and demonstrate the efficiency of using events in comparison to directly analyzing the raw texts. 
    
    \item We propose a generic framework to analyze event gender bias, including
    a calibration technique to offset the potential confounding of gender biases in the event extraction model.
    

\end{itemize}

\section{Experimental Setup} \violet{add a transitional paragraph here to explain that you'll introduce the dataset and models you use for your study.}
In this section, we will introduce our collected corpus and the event extraction model in our study.
\paragraph{Dataset.} 
\violet{give an general overview of the dataset -- although you've mentioned this in the intro, reiterate it here will make the flow more natural.} Our collected corpus contains demographics information and description sections of celebrities from Wikipedia. Table~\ref{table:statistics} shows the statistics of the number of celebrities with \emph{Career} or \emph{Personal Life} sections in our corpora, together with all celebrities we collected. In this work, we only explored celebrities with \emph{Career} or \emph{Personal Life} sections, but there are more sections (e.g., \emph{Politics} and \emph{Background and Family}) in our collected corpus. We encourage interested researchers to further utilize our collected corpus and conduct studies from other perspectives.
In each experiment, we select the same number of female and male celebrities from one occupation for a fair comparison.

\begin{table}[t]
\small
\centering
\resizebox{0.48\textwidth}{!}{%
\begin{tabular}{@{}lllllll@{}}
\toprule
\multicolumn{1}{c}{} & \multicolumn{2}{c}{Career} & \multicolumn{2}{c}{Personal Life} & \multicolumn{2}{c}{Collected} \\ \midrule
Occ & F & M & F & M & F & M \\ \midrule
Acting & 464 & 469 & 464 & 469 & 464 & 469 \\
Writer & 455 & 611 & 319 & 347 & 1,372 & 2,466 \\
Comedian & 380 & 655 & 298 & 510 & 642 & 1,200 \\
Artist & 193 & 30 & 60 & 18 & 701 & 100 \\
Chef & 81 & 141 & 72 & 95 & 176 & 350 \\
Dancer & 334 & 167 & 286 & 127 & 812 & 465 \\
Podcaster & 87 & 183 & 83 & 182 & 149 & 361 \\
Musician & 39 & 136 & 21 & 78 & 136 & 549 \\ \midrule
All & \multicolumn{2}{c}{4,425} & \multicolumn{2}{c}{3,429} & \multicolumn{2}{c}{10,412} \\ \bottomrule
\end{tabular}
}
\caption{Statistics showing the number of celebrities with \emph{Career} section or \emph{Personal Life} section, together with all celebrities we collected. Not all celebrities have \emph{Career} or \emph{Personal Life} sections. } 
\label{table:statistics}
\end{table}

\paragraph{Event Extraction.} There are two definitions of events: one defines an event as the trigger word (usually a verb)~\cite{timebank}, the other defines an event as a complex structure including a trigger, arguments, time, and location~\cite{broad}.\violet{add citations} The corpus following the former definition usually has much broader coverage, while the latter can provide richer information.  
For broader coverage, we choose a state-of-the-art event detection model that focuses on detecting event trigger words by~\citet{Han2019JointEA}.\footnote{We use the code at \url{https://github.com/rujunhan/EMNLP-2019} and reproduce the model trained on the TB-Dense dataset.} 
We use the model trained on the TB-Dense dataset \cite{pustejovsky2003timeml} for two reasons: 1) the model performs better on the TB-Dense dataset; 2) the annotation of the TB-Dense dataset is from the news articles, and it is also where the most content of Wikipedia comes from.\footnote{According to \citet{fetahu2015automated}, more than 20\% of the references are news articles on Wikipedia.} We extract and lemmatize events $e$ from the corpora and count their frequencies $|e|$. Then, we separately construct dictionaries $\mathcal{E}^m = \{e_1^m:|e_1^m|, ..., e_M^m:|e_M^m|\}$ and $\mathcal{E}^f = \{e_1^f:|e_1^f|,  ..., e_F^f:|e_F^f|\}$ mapping events to their frequency for male and female respectively. 

\paragraph{Event Extraction Quality.} To check the model performance on our corpora, we manually annotated events in 10,508 sentences (female: 5,543, male: 4,965) from the Wikipedia corpus. Table~\ref{table:manual} shows that the model performs comparably on our corpora as on the TB-Dense test set. 

\begin{table}[t]
\centering
\begin{tabular}{l l l  l l l}
\toprule
\textbf{Metric} & \textbf{TB-D} & \textbf{S} &\textbf{S-F} & \textbf{S-M} &  \\ \midrule
Precision &    89.2  & 93.5 & 95.3 & 93.4  \\ 
Recall  &  92.6 & 89.8 & 87.1 & 89.8  \\
F1 &  90.9 & 91.6 & 91.0 & 91.6 \\    
\bottomrule
\end{tabular}
\caption{The performance for off-the-shelf event extraction model in both common event extraction dataset TB-Dense (TB-D) and our corpus with manual annotation. \texttt{S} represents the sampled data from the corpus. \texttt{S-F} and \texttt{S-M} represent the sampled data for female career description and male career description separately.} 
\label{table:manual}
\end{table}

\begin{table*}[t]
\small
\centering
\begin{tabular}{l | l | l | l | l}
\toprule
Occupation  & Events in Female Career Description & Events in Male Career Description & WEAT${*}$    & WEAT     \\   \midrule

Writer & \begin{tabular}[c]{@{}l@{}}  \marksymbol{diamond*}{magenta}{\color{magenta}\ divorce}, \marksymbol{diamond*}{magenta}{\color{magenta}\ marriage}, involve,  organize,\\ \marksymbol{diamond*}{magenta}{\color{magenta}\ wedding} \end{tabular}   &  \begin{tabular}[c]{@{}l@{}} argue, \marksymbol{oplus}{violet}{\color{violet} \ election},
\marksymbol{triangle*}{emerald}{\color{emerald}\ protest}, rise, \\ shoot\end{tabular} 
& -0.17 & 1.51\\
\midrule

Acting & \begin{tabular}[c]{@{}l@{}} \marksymbol{diamond*}{magenta}{\color{magenta}\ divorce}, \marksymbol{diamond*}{magenta}{\color{magenta}\ wedding}, guest, name, commit \end{tabular}   &  \begin{tabular}[c]{@{}l@{}} support, \marksymbol{triangle*}{emerald}{\color{emerald}\ arrest}, \marksymbol{triangle*}{emerald}{\color{emerald}\ war}, \marksymbol{square*}{cyan}{\color{cyan} \ sue}, trial\end{tabular}  
& -0.19 & 0.88 \\
\midrule

Comedian & \begin{tabular}[c]{@{}l@{}}  \marksymbol{diamond*}{magenta}{\color{magenta}\ birth},  eliminate, \marksymbol{diamond*}{magenta}{\color{magenta}\ wedding}, \marksymbol{heart}{olive}{\color{olive} \ relocate}, \\ partner \end{tabular} &
\begin{tabular}[c]{@{}l@{}} enjoy, hear, cause, \marksymbol{*}{orange}{\color{orange} \ buy}, conceive\end{tabular}  
& -0.19 & 0.54 \\ 
\midrule

Podcaster &\begin{tabular}[c]{@{}l@{}} \marksymbol{heart}{olive}{\color{olive} \ land}, interview, portray,  \marksymbol{diamond*}{magenta}{\color{magenta}\ married}, report \end{tabular} & 
\begin{tabular}[c]{@{}l@{}} direct, ask, provide, continue, bring  \end{tabular}
& -0.24& 0.53 \\ \midrule

Dancer & \begin{tabular}[c]{@{}l@{}}  \marksymbol{diamond*}{magenta}{\color{magenta}\ married},  \marksymbol{diamond*}{magenta}{\color{magenta}\ marriage},   \marksymbol{heart}{olive}{\color{olive} \ depart}, \marksymbol{heart}{olive}{\color{olive} \ arrive}, \\ organize \end{tabular} &
\begin{tabular}[c]{@{}l@{}} drop, team, choreograph, explore \\ break \end{tabular} 
& -0.14 & 0.22\\ 
\midrule

  Artist &\begin{tabular}[c]{@{}l@{}} paint, exhibit, include, \marksymbol{heart}{olive}{\color{olive} \ return}, teach  \end{tabular} & 
\begin{tabular}[c]{@{}l@{}} start, found, feature, award, begin \end{tabular}
& -0.02 & 0.17 \\ \midrule

Chef &\begin{tabular}[c]{@{}l@{}} \marksymbol{oplus}{violet}{\color{violet} \ hire}, \marksymbol{triangle}{darkgray}{\color{darkgray} \ meet}, debut, eliminate, sign  \end{tabular} & 
\begin{tabular}[c]{@{}l@{}} include, focus, explore, award,  \marksymbol{*}{orange}{\color{orange} \ raise}
\end{tabular}
 & -0.13 & -0.38 \\ \midrule

Musician &\begin{tabular}[c]{@{}l@{}} run, record, \marksymbol{diamond*}{magenta}{\color{magenta}\ death}, found, contribute  \end{tabular} & 
\begin{tabular}[c]{@{}l@{}} sign, direct, produce, premier, open
\end{tabular}
& -0.19 & --0.41 \\ \midrule

\multicolumn{4}{c}{ Annotations:
\marksymbol{diamond*}{magenta}{\color{magenta}\ Life} \
\marksymbol{heart}{olive}{\color{olive} \ Transportation}\
\marksymbol{oplus}{violet}{\color{violet}\ Personell} \
\marksymbol{triangle*}{emerald}{\color{emerald}\ Conflict} \
\marksymbol{square*}{cyan}{\color{cyan} \ Justice} \
\marksymbol{*}{orange}{\color{orange} \ Transaction} \
\marksymbol{triangle}{darkgray}{\color{darkgray} \ Contact} \
}
\\
\bottomrule
\end{tabular}
\caption[Caption]{Top 5 extracted events that occur more often for females and males in \emph{Career} sections across 8 occupations. 
We predict event types by applying EventPlus~\cite{ma2021eventplus} on sentences that contain target events and take the majority vote of the predicted types.  
The event types are from the ACE dataset.\protect\footnotemark\xspace
We calculate WEAT scores with all tokens excluding stop words (WEAT$^{*}$ column) and only detected events (WEAT column) for \emph{Career} sections.
}
\label{tab:result}
\end{table*} 
\begin{figure*}[t]
        \centering
        \subfloat[Male Writers]  {
                \includegraphics[width=0.48\columnwidth, height=0.32\columnwidth]{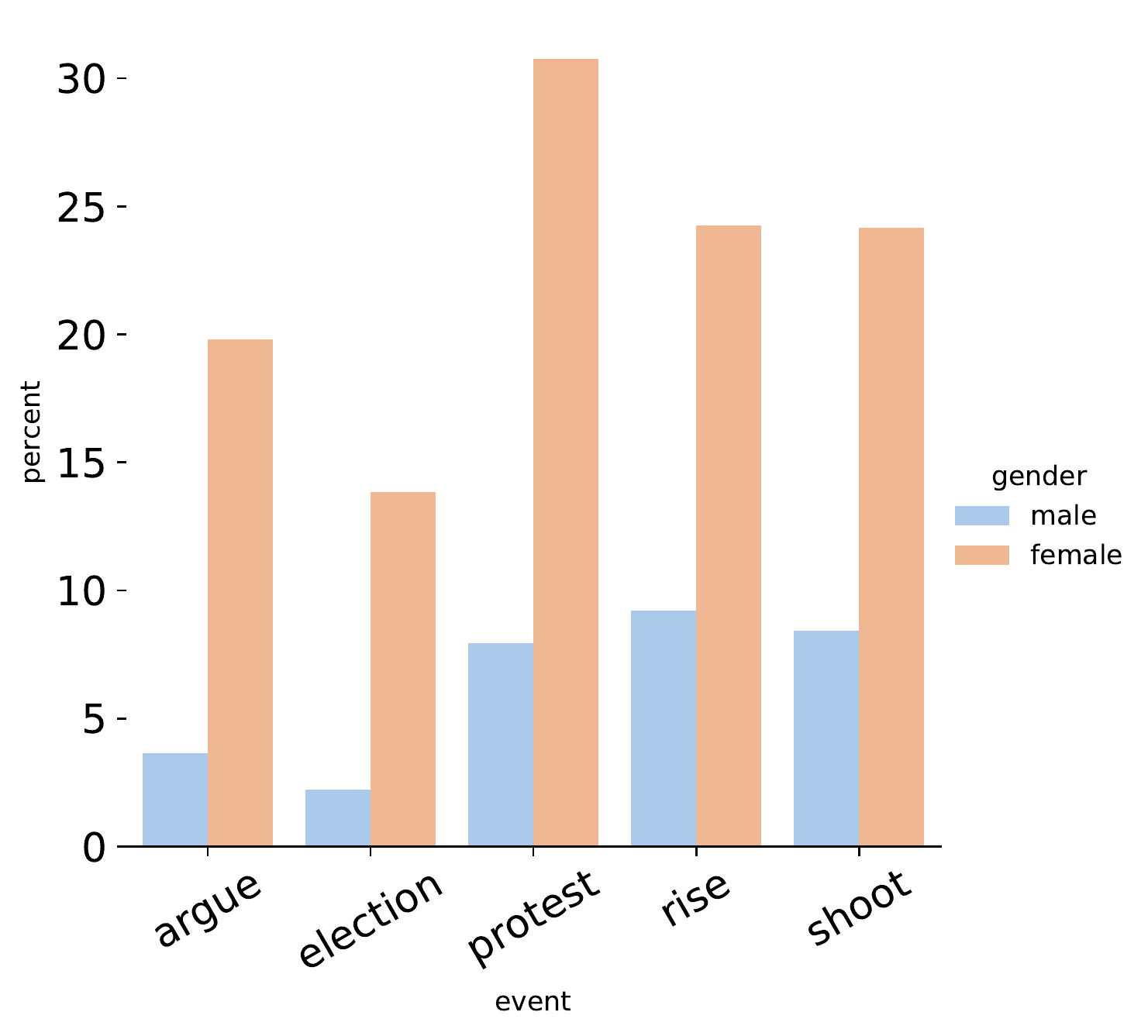}
                \label{fig:male_writer}
        }
        \subfloat[Female Writers] {
                \includegraphics[width=0.48\columnwidth, height=0.32\columnwidth]{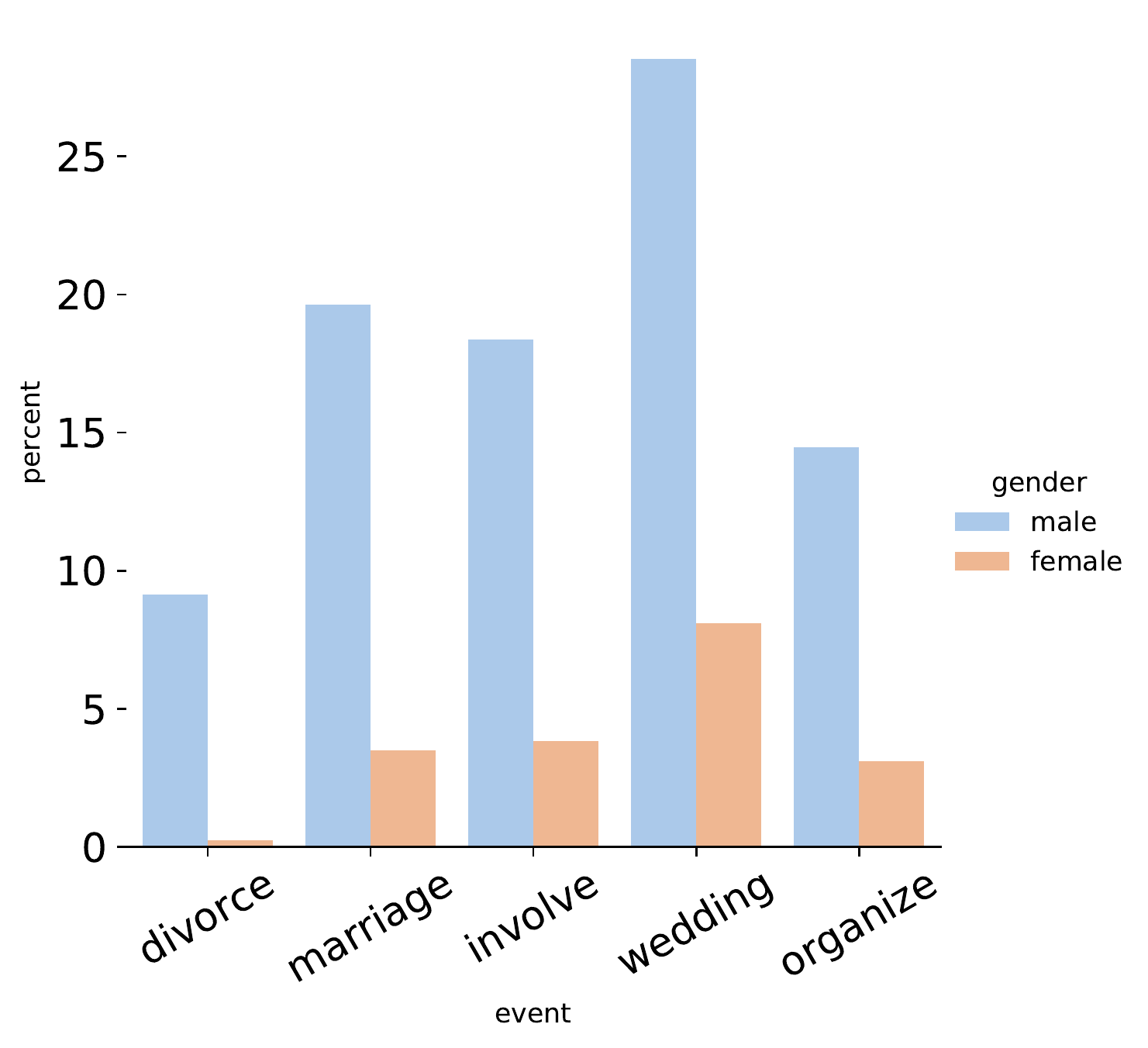}
                \label{fig:female_writer}
            }
        \subfloat[Actor ]  {
                \includegraphics[width=0.48\columnwidth, height=0.32\columnwidth]{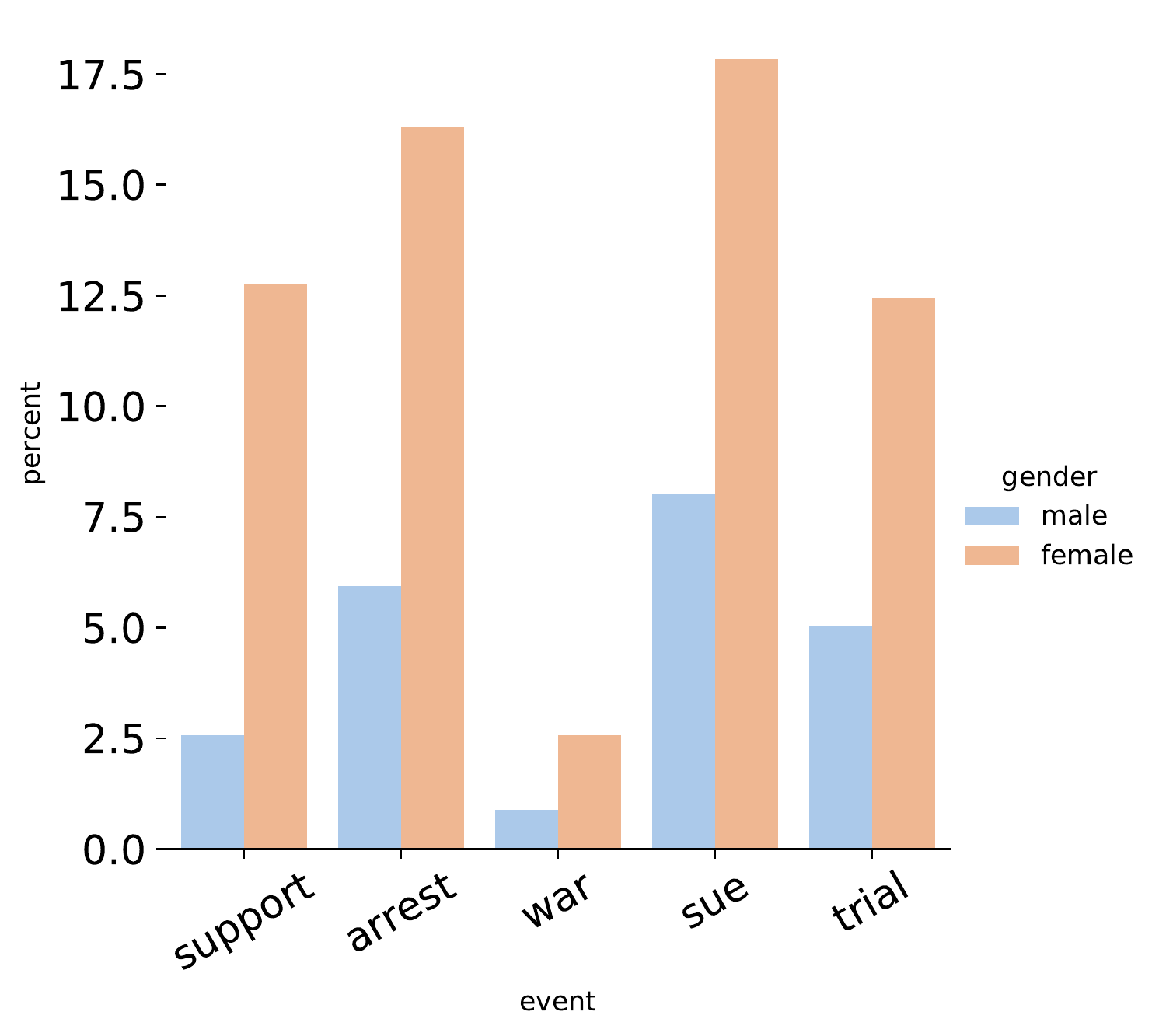}
                \label{fig:actor}
        }
        \subfloat[Actress] {
                \includegraphics[width=0.48\columnwidth, height=0.32\columnwidth]{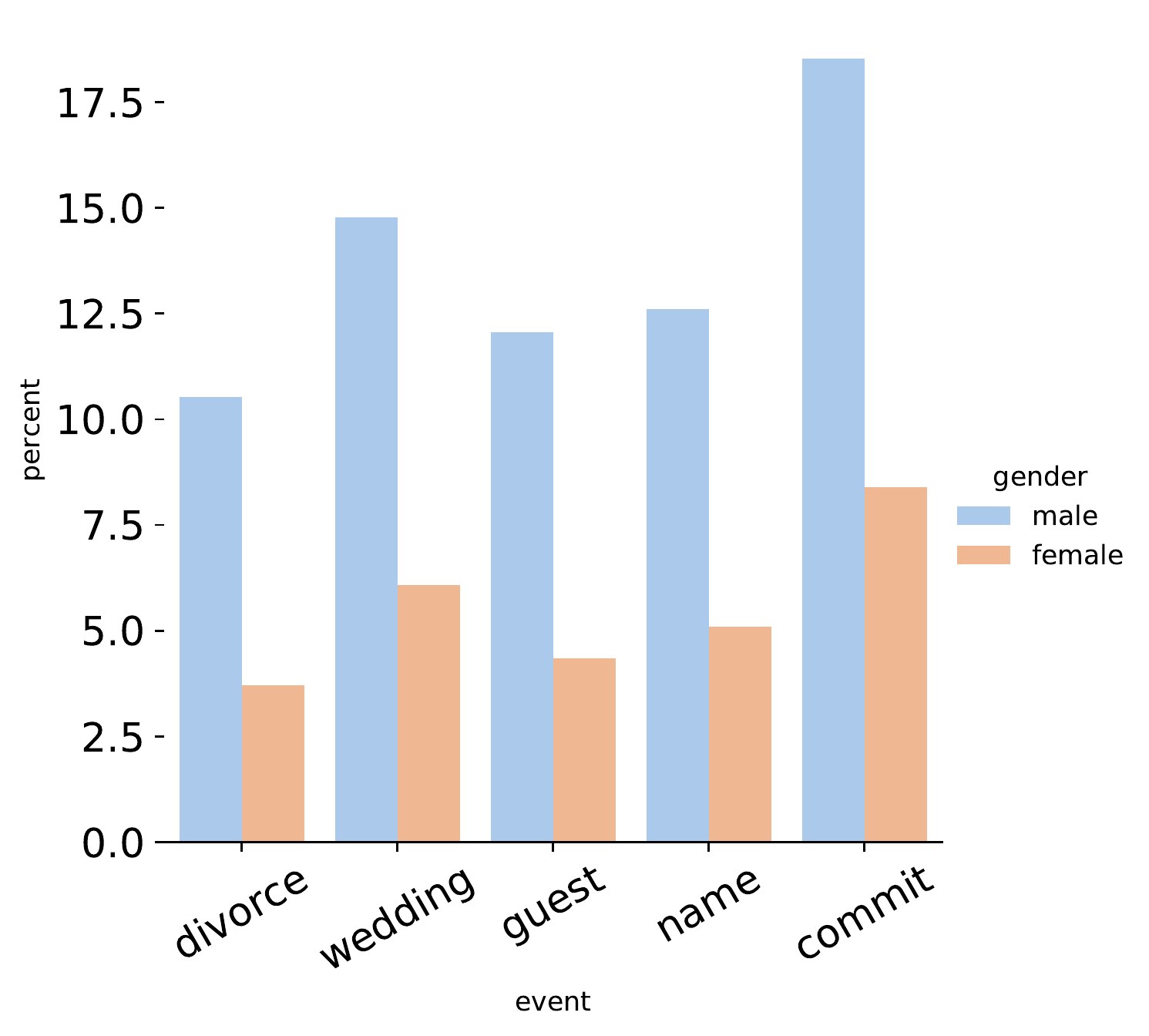}
                \label{fig:actress}
            }
        \caption{The percentile of extracted events among all detected events, sorted by their frequencies in descending order. The smaller the percentile is, the more frequent the event appears in the text. The extracted events are among the top 10\% for the corresponding gender (e.g., extracted female events among all detected events for female writers) and within top 40\% percent for the opposite gender (e.g., extracted female events among all detected events for male writers).  The figure shows that we are not picking rarely-occurred events, and the result is significant. }
        \label{fig:percent}
\end{figure*}

\section{Detecting Gender Biases in Events}
\label{sec:detect}
\paragraph{Odds Ratio.} After applying the event detection model,
we get two dictionaries $\mathcal{E}^m$ and $\mathcal{E}^f$ that have events as keys and their corresponding occurrence frequencies as values. Among all events, we focus on those with distinct occurrences in males and females descriptions (e.g., \texttt{work} often occurs at a similar frequency for both females and males in \emph{Career} sections, and we thus neglect it from our analysis). 
We use the Odds Ratio (OR)~\cite{Szumilas2010ExplainingOR} to find the events with large frequency differences for females and males, which indicates that they might potentially manifest gender biases. 
For an event $e_n$, we calculate its odds ratio as the odds of having it in the male event list divided by the odds of having it in the female event list:
\begin{equation}
\frac{\mathcal{E}^m(e_n)}{\sum_{\substack{e_i^m\ne e_n \\ i \in [1, ..., M]}}^{i} \mathcal{E}^m(e_i^m)} / \frac{\mathcal{E}^f(e_n)}{\sum_{\substack{e_j^f\ne e_n\\j \in [1,..., F]}}^{j} \mathcal{E}^f(e_j^f)}
\label{equ:or}
\end{equation}

The larger the OR is, the more likely an event will occur in male than female sections by  Equation~\ref{equ:or}. After obtaining a list of events and their corresponding OR, 
we sort the events by OR in descending order. The top $k$ events are more likely to appear for males and the last $k$ events for females.


\paragraph{Calibration.} 
The difference of event frequencies might come from the model bias, as shown in other tasks (e.g., gender bias in coreference resolution model~\cite{Zhao2018GenderBI}). To offset the potential confounding that could be brought by the event extraction model and estimate the actual event frequency, we propose a calibration strategy by 1) generating data that contains target events; 2) testing the model performance for females and males separately in the generated data, 3) and using the model performance to estimate real event occurrence frequencies. 

\footnotetext{ACE dataset: \url{https://www.ldc.upenn.edu/collaborations/past-projects/ace}}

We aim to calibrate the top 50 most skewed events in females' and males' \emph{Career} and \emph{Personal Life} descriptions after using the OR separately. First, we follow two steps to generate a synthetic dataset:
\begin{enumerate}
    \item For each target event, we select all sentences where the model successfully detected the target event. 
    For each sentence, we manually verify the correctness of the extracted event and discard the incorrect ones. For the rest, we use the verified sentences to create more ground truth; we call them \emph{template sentences}. 
    
    \item  For each template sentence, we find the celebrity's first name 
    and mark it as a \texttt{Name Placeholder}, then we replace it with 50 female names and 50 male names that are sampled from the name list by \citet{checklist}. If the gender changes during the name replacement (e.g., Mike to Emily), we replace the corresponding pronouns (e.g., he to she) and gender attributes~\cite{Zhao2018GenderBI} (e.g., Mr to Miss) in the template sentences. As a result, we get 100 data points for each template sentence with automatic annotations. If there is no first name in the sentence, we replace the pronouns and gender attributes.
\end{enumerate}

\footnotetext{We did not show the result for the \texttt{artists} and \texttt{musicians} due to the small data size.}

After getting the synthetic data, we run the event extraction model again. We use the detection recall among the generated instances to calibrate the frequency $|e|$ for each target event and estimate the actual frequency $|e|^{*}$, following: 
\begin{equation}
    |e|^{*} = \frac{|e|}{TP(e) / (TP(e) + FP(e))}
\label{equ:recall}
\end{equation}

 Then, we replace $|e|$ with $|e|^{*}$ in Equation~\ref{equ:or}, and get $k$ female and $k$ male events by sorting OR as before.  Note that we observe the model performances are mostly unbiased, and we have only calibrated events that have different performances for females and males over a threshold (i.e., 0.05).\footnote{ Calibration details and quantitative result in App.~\ref{appendix:cali}.} 



\paragraph{WEAT score.} We further check if the extracted events are associated with gender attributes (e.g., \texttt{she and her} for females, and \texttt{he and him} for males) in popular neural word embeddings like Glove~\cite{glove}. We quantify this with the Word Embedding Association Test (WEAT) ~\cite{WEAT}, a popular method for measuring biases in text. 
Intuitively, WEAT takes a list of tokens that represent a concept (in our case, \emph{extracted events}) and verifies whether these tokens have a shorter distance towards female attributes or male attributes. 
A positive value of WEAT score indicates that female events are closer to female attributes, and male events are closer to male attributes\violet{and/or male events are closer to male attributes?} in the word embedding, while a negative value indicates that female events are closer to male attributes and vice versa.\footnote{Details of WEAT score experiment in App.~\ref{appendix:weat}.}

To show the effectiveness of using events as a lens for gender bias analysis, we compute WEAT scores on the raw texts and detected events separately. For the former, we take all tokens excluding stop words.\footnote{We use spaCy (\url{https://spacy.io/}) to tokenize the corpus and remove stop words.} Together with gender attributes from ~\citet{WEAT}, we calculate and show the WEAT scores under two settings as ``WEAT$^{*}$'' for the raw texts and  ``WEAT'' for the detected events. 

\begin{table*}[h!]
\resizebox{\textwidth}{!}{
\small
\centering
\begin{tabular}{l | l | l | l | l}
\toprule
Occupation  & Events in Female Personal Life Description & Events in Male Personal Life Description & WEAT${*}$    & WEAT     \\   \midrule

Writer 
& 
\begin{tabular}[c]{@{}l@{}} 
bury, \marksymbol{diamond*}{magenta}{\color{magenta}\ birth}, attend,  \marksymbol{triangle*}{emerald}{\color{emerald} \ war},  grow \end{tabular}   
&  
\begin{tabular}[c]{@{}l@{}} know, report,
come, \marksymbol{square*}{cyan}{\color{cyan} \ charge}, publish\end{tabular} 
& -0.05 & 0.31\\
\midrule

Acting 
&
\begin{tabular}[c]{@{}l@{}} \marksymbol{diamond*}{magenta}{\color{magenta}\ pregnant}, practice, wedding, record, convert \end{tabular}   
&
\begin{tabular}[c]{@{}l@{}} accuse, \marksymbol{heart}{olive}{\color{olive} \ trip}, \marksymbol{heart}{olive}{\color{olive} \ fly}, \marksymbol{triangle*}{emerald}{\color{emerald} \ assault}, endorse\end{tabular}  
& -0.14 & 0.54 \\
\midrule

Comedian 
&
\begin{tabular}[c]{@{}l@{}} feel,  \marksymbol{diamond*}{magenta}{\color{magenta}\ birth}, fall, open, decide \end{tabular} 
&
\begin{tabular}[c]{@{}l@{}} \marksymbol{heart}{olive}{\color{olive} \  visit}, create, spend, propose, lawsuit\end{tabular}  
& -0.07 & 0.07 \\ 
\midrule

Podcaster &\begin{tabular}[c]{@{}l@{}}  date, describe, tell, life, come \end{tabular} 
& 
\begin{tabular}[c]{@{}l@{}} play, write, \marksymbol{diamond*}{magenta}{\color{magenta} \ born}, release, claim  \end{tabular}
& -0.13 & 0.57 \\ \midrule

Dancer 
& \begin{tabular}[c]{@{}l@{}}  \marksymbol{diamond*}{magenta}{\color{magenta}\ marry},  describe,   diagnose,  expect, speak \end{tabular} 
&
\begin{tabular}[c]{@{}l@{}} hold, involve, \marksymbol{*}{orange}{\color{orange} \ award}, run, serve \end{tabular} 
& -0.03 & 0.41\\ 
\midrule

Chef 
&
\begin{tabular}[c]{@{}l@{}} \marksymbol{diamond*}{magenta}{\color{magenta} \ death}, serve, announce, describe, \marksymbol{diamond*}{magenta}{\color{magenta}\ born}  \end{tabular} 
&
\begin{tabular}[c]{@{}l@{}} \marksymbol{diamond*}{magenta}{\color{magenta}\ birth}, lose, \marksymbol{diamond*}{magenta}{\color{magenta}\ divorce}, speak,  \ \marksymbol{triangle}{darkgray}{\color{darkgray} \ meet}
\end{tabular}
 & -0.02 & -0.80 \\ \midrule

\multicolumn{4}{c}{ Annotations:
\marksymbol{diamond*}{magenta}{\color{magenta}\ Life} \
\marksymbol{heart}{olive}{\color{olive} \ Transportation}\
\marksymbol{oplus}{violet}{\color{violet}\ Personell} \
\marksymbol{triangle*}{emerald}{\color{emerald}\ Conflict} \
\marksymbol{square*}{cyan}{\color{cyan} \ Justice} \
\marksymbol{*}{orange}{\color{orange} \ Transaction} \
\marksymbol{triangle}{darkgray}{\color{darkgray} \ Contact} \
}
\\
\bottomrule
\end{tabular}
}
\caption[Caption]{Top 5 events in \emph{Personal Life} section across 6 occupations.\protect\footnotemark\xspace There are more \texttt{Life} events (e.g., ``birth'' and ``marry'') in females' personal life descriptions than males' for most occupations. While for males, although we see more life-related events than in the \emph{Career} section, there are events like ``awards'' even in the \emph{Personal Life} section. 
The findings further show our work is imperative and addresses the importance of not intermingling the professional career with personal life regardless of gender during the future editing on Wikipedia.
}

\label{table:pl_result}
\end{table*}

\section{Results} 
\paragraph{The Effectiveness of our Analysis Framework.} Table~\ref{tab:result} and Table~\ref{table:pl_result} show the associations of both raw texts and the extracted events in \emph{Career} and \emph{Personal Life} sections for females and males across occupations after the calibration.
The values in WEAT$^{*}$ columns in both tables indicate that there was only a weak association of words in raw texts with gender. In contrast, the extracted events are associated with gender for most occupations. It shows the effectiveness of the event extraction model and our analysis method.

\paragraph{The Significance of the Analysis Result.}
There is a possibility that our analysis, although it picks out distinct events for different genders, identifies the events that are infrequent for all genders and that the frequent events have similar distributions across genders. 
To verify, we sort all detected events from our corpus by frequencies in descending order. Then, we calculate the percentile of extracted events in the sorted list. The smaller the percentile is, the more frequent the event appears in the text. Figure~\ref{fig:percent} shows that we are not picking the events that rarely occur, which shows the significance of our result.\footnote{See plots for all occupations in Appendix~\ref{appendix:distribution}.} For example, Figure~\ref{fig:male_writer} and Figure~\ref{fig:female_writer} show the percentile of frequencies for selected male and female events among all events frequencies in the descending order for male and female writers, respectively. We can see that for the corresponding gender, event frequencies are among the top 10\%. These events occur less frequently for the opposite gender but still among the top 40\%. 
\paragraph{Findings and Discussions.} We find that there are more \texttt{Life} events for females than males in both \emph{Career} and \emph{Personal Life} sections. On the other hand, for males, there are events like ``awards''  even in their \emph{Personal Life} section. The mixture of personal life with females' professional career events and career achievements with males' personal life events carries implicit gender bias and reinforces the gender stereotype. It potentially leads to career, marital, and parental status discrimination towards genders and jeopardizes gender equality in society. We recommend: 1) Wikipedia editors to restructure pages to ensure that personal life-related events (e.g., marriage and divorce) are written in the \emph{Personal Life} section, and professional events (e.g., award) are written in \emph{Career} sections regardless of gender; 2) future contributors should also be cautious and not intermingle \emph{Personal Life} and \emph{Career} when creating the Wikipedia pages from the start.
\section{Conclusion}
We conduct the first event-centric gender bias analysis at the corpus level and compose a corpus by scraping Wikipedia to facilitate the study. 
Our analysis discovers that the collected corpus has event gender biases. For example, personal life related events (e.g., \texttt{marriage}) are more likely to appear for females than males even in \emph{Career} sections. 
We hope our work brings awareness of potential gender biases in knowledge sources such as Wikipedia, and urges Wikipedia editors and contributors to be cautious when contributing to the pages. 




\section*{Acknowledgments}
This material is based on research supported by IARPA BETTER program via Contract No. 2019-19051600007. The views and conclusions contained herein are those of the authors and should not be interpreted as necessarily representing the official policies or endorsements, either expressed or implied, of IARPA or the U.S. Government.

\section*{Ethical Considerations}
Our corpus is collected from Wikipedia. The content of personal life description, career description, and demographic information is all public to the general audience. Note that our collected corpus might be used for malicious purposes. For example, it can serve as a source by text generation tools to generate text highlighting gender stereotypes.

This work is subject to several limitations: First, it is important to understand and analyze the event gender bias for gender minorities, missing from our work because of scarce resources online. Future research can build upon our work, go beyond the binary gender and incorporate more analysis. Second, our study focuses on the Wikipedia pages for celebrities for two additional reasons besides the broad impact of Wikipedia: 1) celebrities' Wikipedia pages are more accessible than non-celebrities. Our collected Wikipedia pages span across 8 occupations to increase the representation of our study; 2) Wikipedia contributors have been extensively updating celebrities' Wikipedia pages every day. Wikipedia develops at a rate of over 1.9 edits every second, performed by editors from all over the world~\cite{wikistats}. The celebrities' pages get more attention and edits, thus better present how the general audience perceives important information and largely reduce the potential biases that could be introduced in personal writings. Please note that although we try to make our study as representative as possible, it cannot represent certain groups or individuals' perceptions.

Our model is trained on TB-Dense, a public dataset coming from news articles. These do not contain any explicit detail that leaks information about a user's name, health, negative financial status, racial or ethnic origin, religious or philosophical affiliation or beliefs, trade union membership, alleged or actual crime commission.


\bibliographystyle{acl_natbib}
\bibliography{acl2020}

\begin{thebibliography}{20}
\expandafter\ifx\csname natexlab\endcsname\relax\def\natexlab#1{#1}\fi

\bibitem[{wik(2021)}]{wikistats}
 2021.
\newblock Wikipedia:statistics - wikipedia.
\newblock \url{https://en.wikipedia.org/wiki/Wikipedia:Statistics}.
\newblock (Accessed on 02/01/2021).

\bibitem[{Ahn(2006)}]{broad}
David Ahn. 2006.
\newblock The stages of event extraction.
\newblock In \emph{Proceedings of the Workshop on Annotating and Reasoning
  about Time and Events}, pages 1--8.

\bibitem[{Caliskan et~al.(2017)Caliskan, Bryson, and Narayanan}]{WEAT}
A.~Caliskan, J.~Bryson, and A.~Narayanan. 2017.
\newblock Semantics derived automatically from language corpora contain
  human-like biases.
\newblock \emph{Science}, 356:183 -- 186.

\bibitem[{Fetahu et~al.(2015)Fetahu, Markert, and Anand}]{fetahu2015automated}
Besnik Fetahu, Katja Markert, and Avishek Anand. 2015.
\newblock Automated news suggestions for populating wikipedia entity pages.
\newblock In \emph{Proceedings of the 24th ACM International on Conference on
  Information and Knowledge Management}, pages 323--332.

\bibitem[{Gerken(2010)}]{Gerken2010HowCU}
Joseph~L. Gerken. 2010.
\newblock How courts use wikipedia.
\newblock \emph{The Journal of Appellate Practice and Process}, 11:191.

\bibitem[{Han et~al.(2019)Han, Ning, and Peng}]{Han2019JointEA}
Rujun Han, Qiang Ning, and Nanyun Peng. 2019.
\newblock Joint event and temporal relation extraction with shared
  representations and structured prediction.
\newblock In \emph{Proceedings of the 2019 Conference on Empirical Methods in
  Natural Language Processing and the 9th International Joint Conference on
  Natural Language Processing (EMNLP-IJCNLP)}, pages 434--444, Hong Kong,
  China. Association for Computational Linguistics.

\bibitem[{Jolly et~al.(2014)Jolly, Griffith, Decastro, Stewart, Ubel, and
  Jagsi}]{Jolly2014GenderDI}
S.~Jolly, K.~Griffith, R.~Decastro, A.~Stewart, P.~Ubel, and R.~Jagsi. 2014.
\newblock Gender differences in time spent on parenting and domestic
  responsibilities by high-achieving young physician-researchers.
\newblock \emph{Annals of Internal Medicine}, 160:344--353.

\bibitem[{Kousha and Thelwall(2017)}]{Kousha2017AreWC}
K.~Kousha and M.~Thelwall. 2017.
\newblock Are wikipedia citations important evidence of the impact of scholarly
  articles and books?
\newblock \emph{Journal of the Association for Information Science and
  Technology}, 68.

\bibitem[{Leopold(2018)}]{Leopold2018GenderDI}
T.~Leopold. 2018.
\newblock Gender differences in the consequences of divorce: A study of
  multiple outcomes.
\newblock \emph{Demography}, 55:769 -- 797.

\bibitem[{Ma et~al.(2021)Ma, Sun, Yang, Huang, Wen, Singh, Han, and
  Peng}]{ma2021eventplus}
Mingyu~Derek Ma, Jiao Sun, Mu~Yang, Kung-Hsiang Huang, Nuan Wen, Shikhar Singh,
  Rujun Han, and Nanyun Peng. 2021.
\newblock Eventplus: A temporal event understanding pipeline.
\newblock In \emph{2021 Annual Conference of the North American Chapter of the
  Association for Computational Linguistics (NAACL), Demonstrations Track}.

\bibitem[{Pei and Jurgens(2020)}]{Pei2020QuantifyingII}
Jiaxin Pei and David Jurgens. 2020.
\newblock Quantifying intimacy in language.
\newblock In \emph{Proceedings of the 2020 Conference on Empirical Methods in
  Natural Language Processing (EMNLP)}, pages 5307--5326, Online. Association
  for Computational Linguistics.

\bibitem[{Pennington et~al.(2014)Pennington, Socher, and Manning}]{glove}
Jeffrey Pennington, Richard Socher, and Christopher Manning. 2014.
\newblock {G}lo{V}e: Global vectors for word representation.
\newblock In \emph{Proceedings of the 2014 Conference on Empirical Methods in
  Natural Language Processing ({EMNLP})}, pages 1532--1543, Doha, Qatar.
  Association for Computational Linguistics.

\bibitem[{Pustejovsky et~al.(2003{\natexlab{a}})Pustejovsky, Castano, Ingria,
  Sauri, Gaizauskas, Setzer, Katz, and Radev}]{pustejovsky2003timeml}
James Pustejovsky, Jos{\'e}~M Castano, Robert Ingria, Roser Sauri, Robert~J
  Gaizauskas, Andrea Setzer, Graham Katz, and Dragomir~R Radev.
  2003{\natexlab{a}}.
\newblock Timeml: Robust specification of event and temporal expressions in
  text.
\newblock \emph{New directions in question answering}, 3:28--34.

\bibitem[{Pustejovsky et~al.(2003{\natexlab{b}})Pustejovsky, Hanks, Sauri, See,
  Gaizauskas, Setzer, Radev, Sundheim, Day, Ferro et~al.}]{timebank}
James Pustejovsky, Patrick Hanks, Roser Sauri, Andrew See, Robert Gaizauskas,
  Andrea Setzer, Dragomir Radev, Beth Sundheim, David Day, Lisa Ferro, et~al.
  2003{\natexlab{b}}.
\newblock The timebank corpus.
\newblock In \emph{Corpus linguistics}, volume 2003, page~40. Lancaster, UK.

\bibitem[{Rashkin et~al.(2018)Rashkin, Sap, Allaway, Smith, and
  Choi}]{Rashkin2018Event2MindCI}
Hannah Rashkin, Maarten Sap, Emily Allaway, Noah~A. Smith, and Yejin Choi.
  2018.
\newblock {E}vent2{M}ind: Commonsense inference on events, intents, and
  reactions.
\newblock In \emph{Proceedings of the 56th Annual Meeting of the Association
  for Computational Linguistics (Volume 1: Long Papers)}, pages 463--473,
  Melbourne, Australia. Association for Computational Linguistics.

\bibitem[{Ribeiro et~al.(2020)Ribeiro, Wu, Guestrin, and Singh}]{checklist}
Marco~Tulio Ribeiro, Tongshuang Wu, Carlos Guestrin, and Sameer Singh. 2020.
\newblock Beyond accuracy: Behavioral testing of {NLP} models with
  {C}heck{L}ist.
\newblock In \emph{Proceedings of the 58th Annual Meeting of the Association
  for Computational Linguistics}, pages 4902--4912, Online. Association for
  Computational Linguistics.

\bibitem[{Szumilas(2010)}]{Szumilas2010ExplainingOR}
M.~Szumilas. 2010.
\newblock Explaining odds ratios.
\newblock \emph{Journal of the Canadian Academy of Child and Adolescent
  Psychiatry = Journal de l'Academie canadienne de psychiatrie de l'enfant et
  de l'adolescent}, 19 3:227--9.

\bibitem[{Young et~al.(2016)Young, Wigdor, and Kane}]{Young2016ItsNW}
A.~Young, Ari~D. Wigdor, and Gerald Kane. 2016.
\newblock It's not what you think: Gender bias in information about fortune
  1000 ceos on wikipedia.
\newblock In \emph{ICIS}.

\bibitem[{Yu et~al.(2015)Yu, Gormley, and Dredze}]{ACE}
Mo~Yu, Matthew~R. Gormley, and Mark Dredze. 2015.
\newblock Combining word embeddings and feature embeddings for fine-grained
  relation extraction.
\newblock In \emph{Proceedings of the 2015 Conference of the North {A}merican
  Chapter of the Association for Computational Linguistics: Human Language
  Technologies}, pages 1374--1379, Denver, Colorado. Association for
  Computational Linguistics.

\bibitem[{Zhao et~al.(2018)Zhao, Wang, Yatskar, Ordonez, and
  Chang}]{Zhao2018GenderBI}
Jieyu Zhao, Tianlu Wang, Mark Yatskar, Vicente Ordonez, and Kai-Wei Chang.
  2018.
\newblock Gender bias in coreference resolution: Evaluation and debiasing
  methods.
\newblock In \emph{Proceedings of the 2018 Conference of the North {A}merican
  Chapter of the Association for Computational Linguistics: Human Language
  Technologies, Volume 2 (Short Papers)}, pages 15--20, New Orleans, Louisiana.
  Association for Computational Linguistics.

\end{thebibliography}

\appendix
\clearpage

\section{Appendix}

\subsection{Quality Check: Event Detection Model}
To test the performance of the event extraction model in our collected corpus from Wikipedia. We manually annotated events in 10,508 (female: 5,543, male: 4,965) sampled sentences from the \emph{Career} section in our corpus. Our annotators are two volunteers who are not in the current project but have experience with event detection tasks. We asked annotators to annotate all event trigger words in the text. During annotation, we follow the definition of events from the ACE annotation guideline.\footnote{\url{https://www.ldc.upenn.edu/sites/www.ldc.upenn.edu/files/english-events-guidelines-v5.4.3.pdf}} We use the manual annotation as the ground truth and compare it with the event detection model output to calculate the metrics (i.e., precision, recall, and F1) in Table~\ref{table:manual}.

\subsection{Calibration Details}
\label{appendix:cali}

\begin{figure*}[t]
\centering
\vspace{-1cm}
\subfloat[Writers-\emph{c}]{\label{fig:apd_female_writer} \includegraphics[width=0.23\textwidth, height=0.15\textwidth]{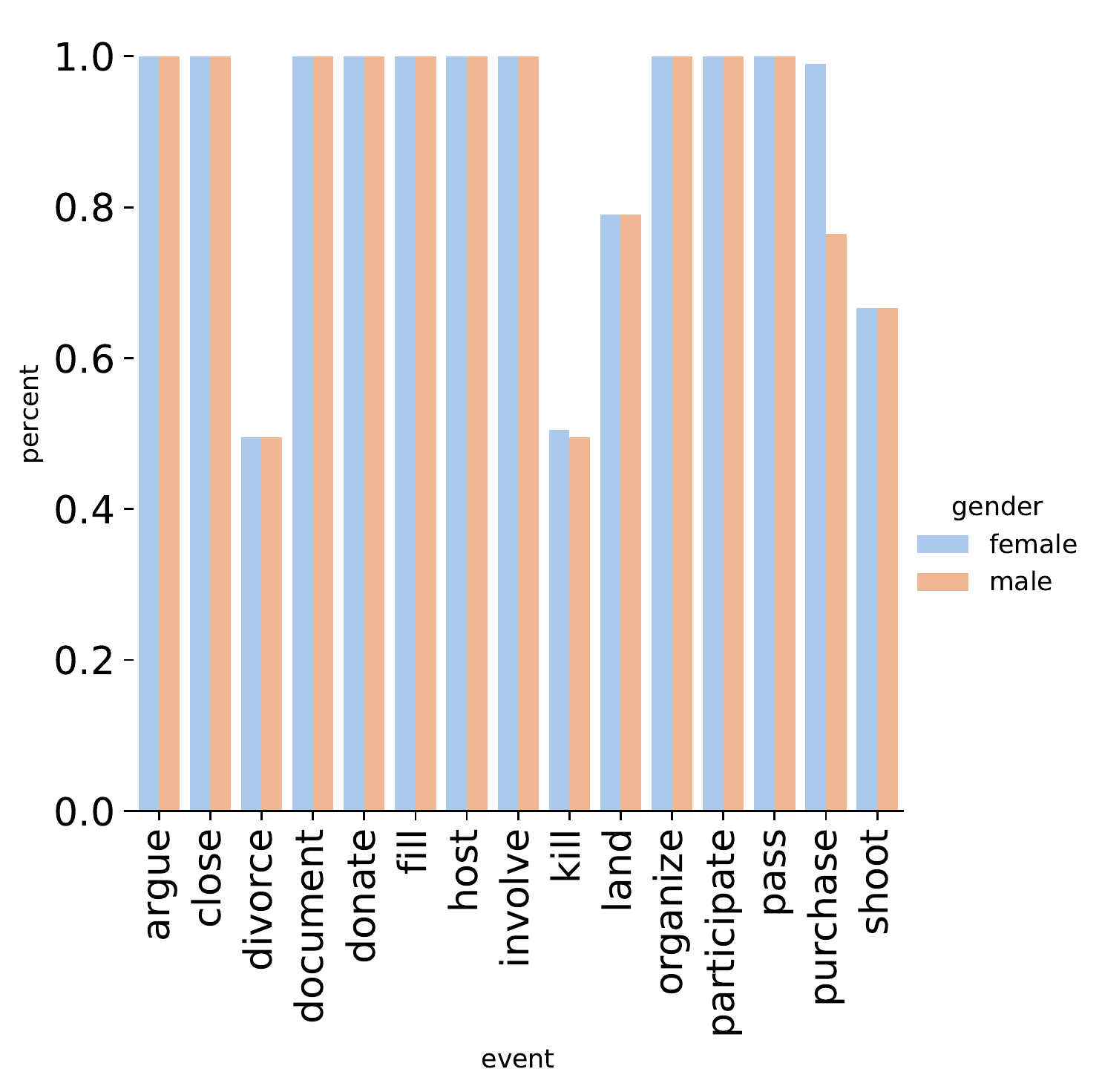}}
\vspace{-0.15cm}
\subfloat[Comedians-\emph{c}]{\label{fig:apd_comedian} \includegraphics[width=0.23\textwidth, height=0.15\textwidth]{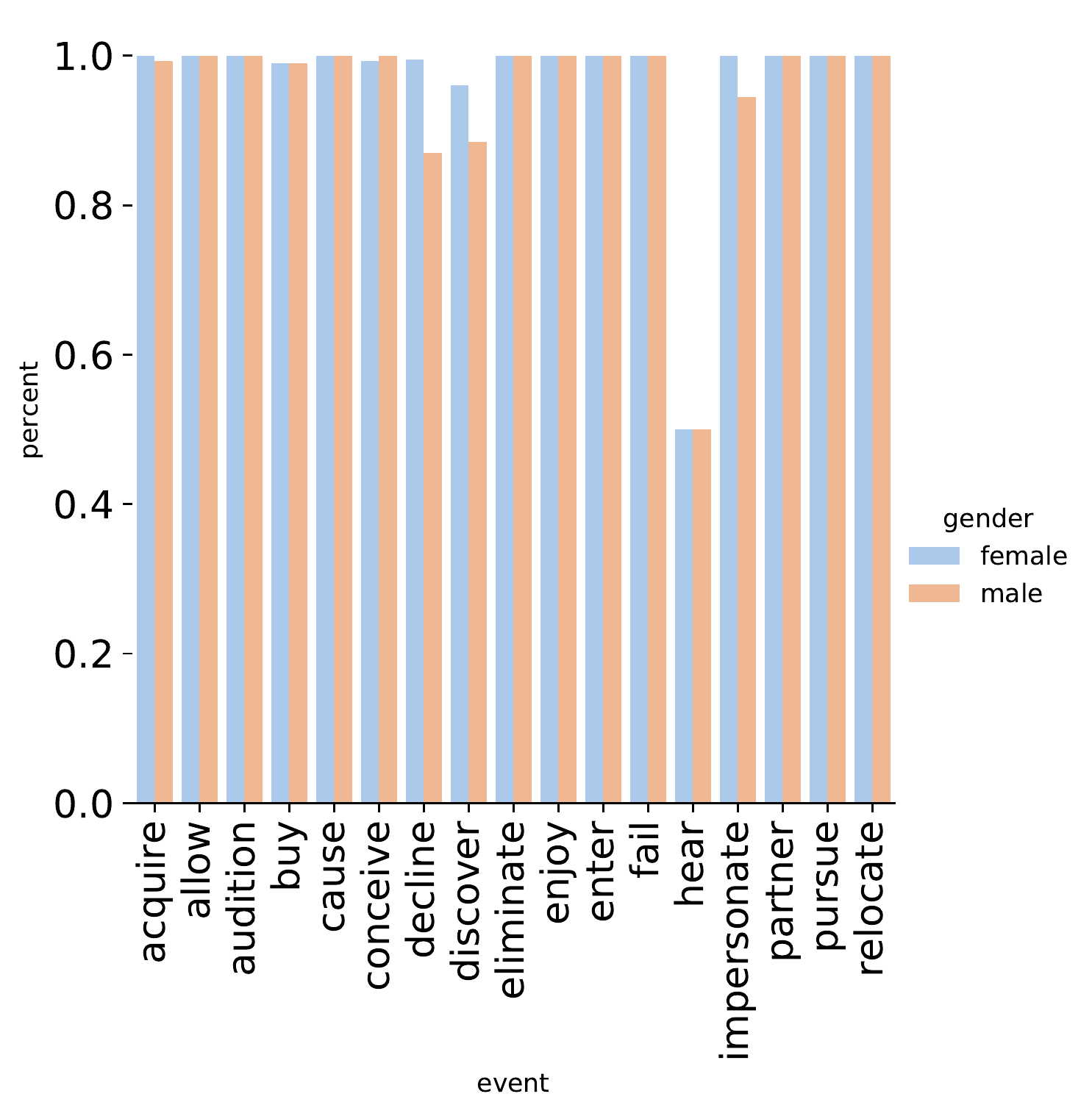}}
\vspace{-0.15cm}
\subfloat[Podcasters-\emph{c}]{\label{fig:apd_podcaster} \includegraphics[width=0.23\textwidth, height=0.15\textwidth]{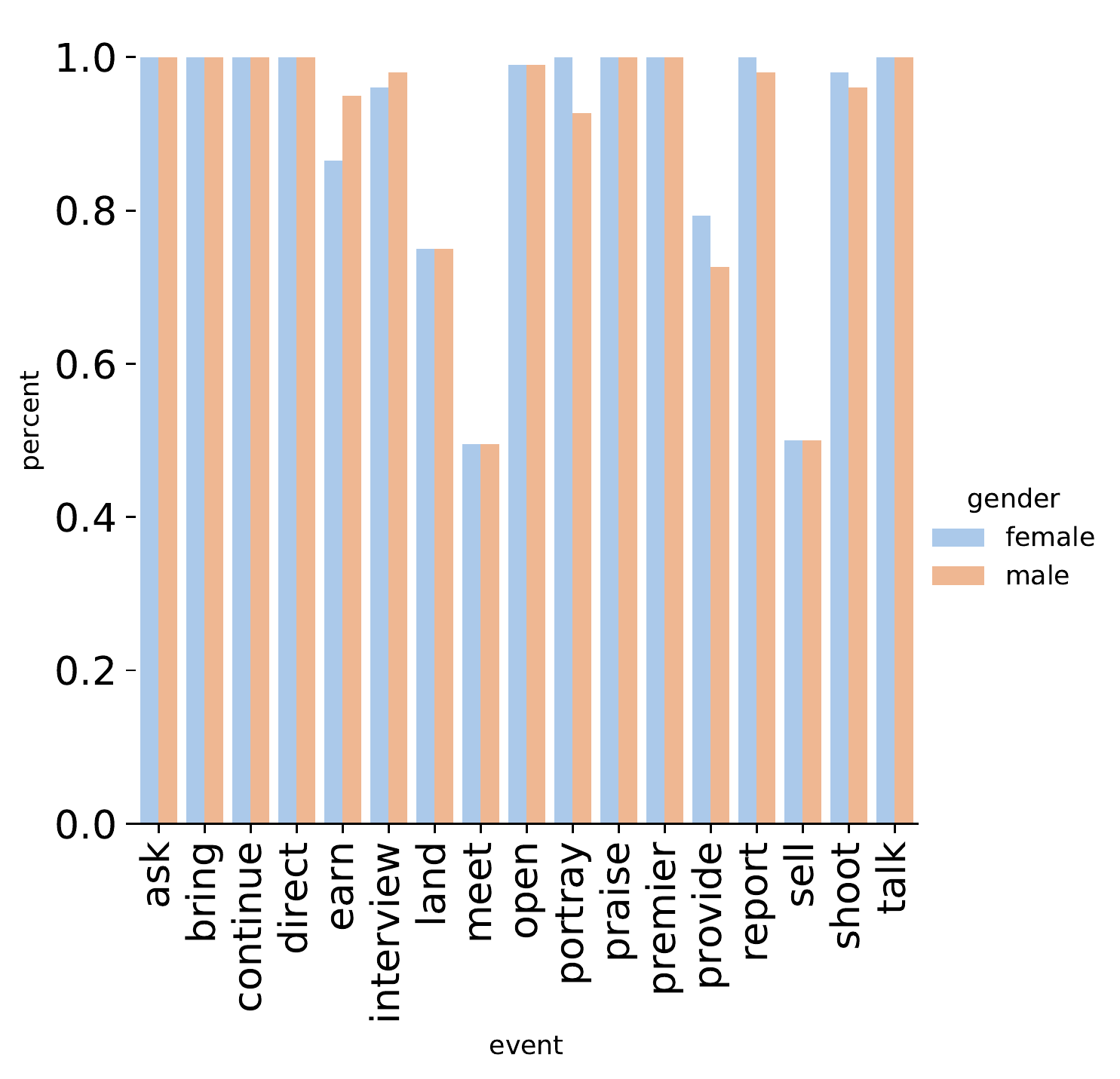}}
\vspace{-0.15cm}
\subfloat[Dancers-\emph{c}]{\label{fig:apd_male_dancer} \includegraphics[width=0.23\textwidth, height=0.15\textwidth]{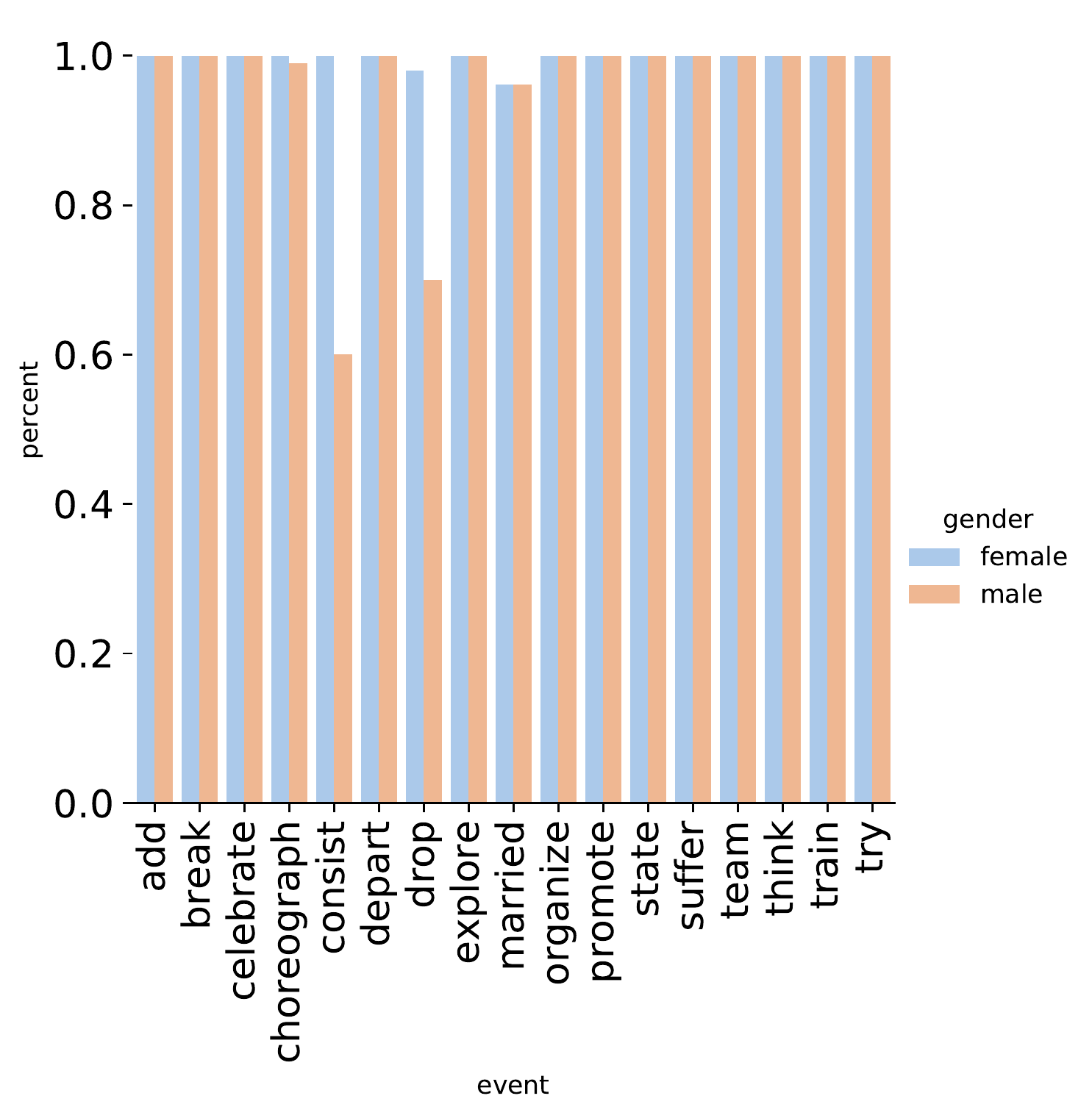}}\\
\vspace{-0.04cm}
\subfloat[Artists-\emph{c}]{\label{fig:apd_female_artists} \includegraphics[width=0.23\textwidth, height=0.15\textwidth]{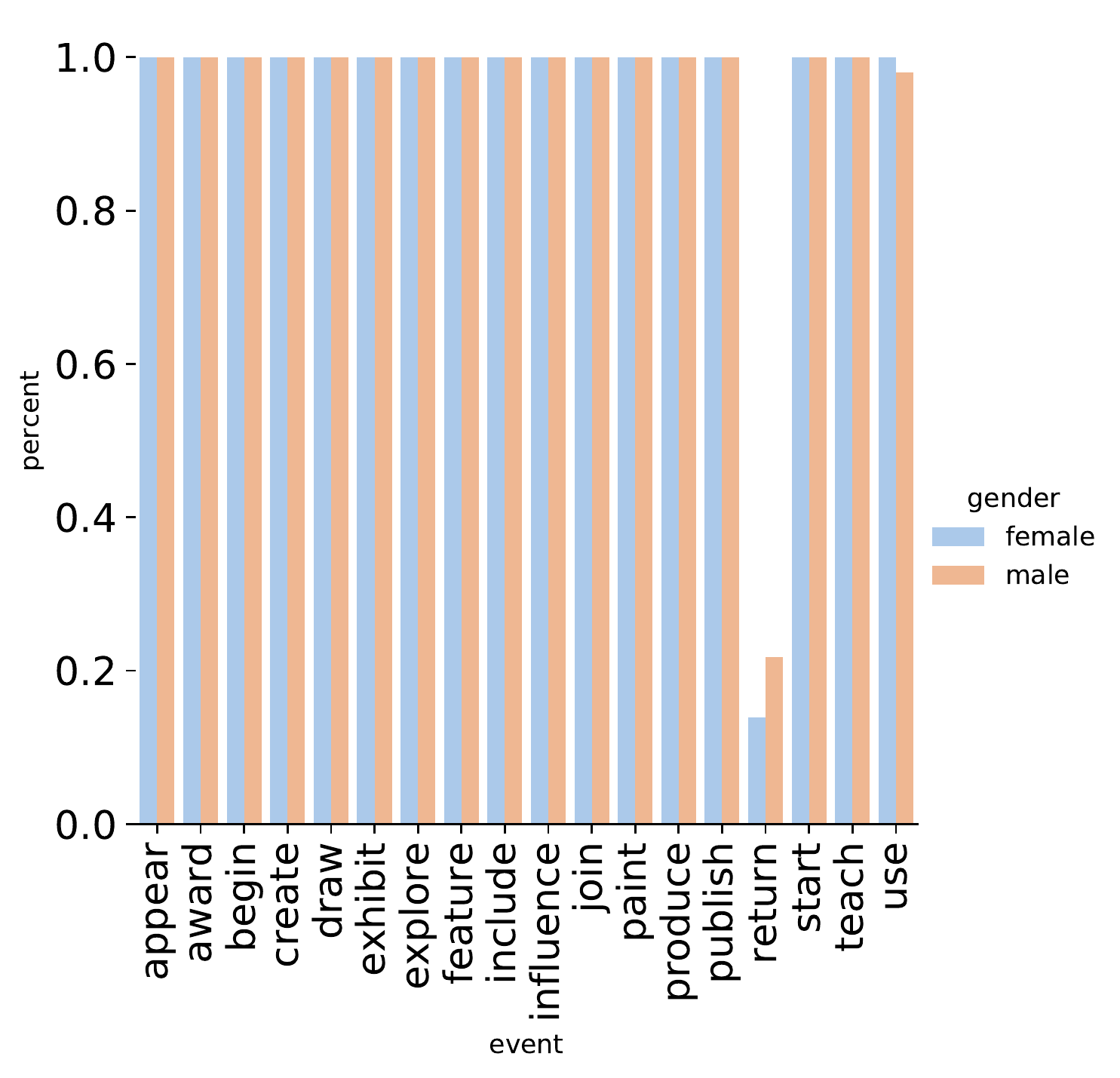}}
\vspace{-0.04cm}
\subfloat[Musicians-\emph{c}]{\label{fig:apd_musician} \includegraphics[width=0.23\textwidth, height=0.15\textwidth]{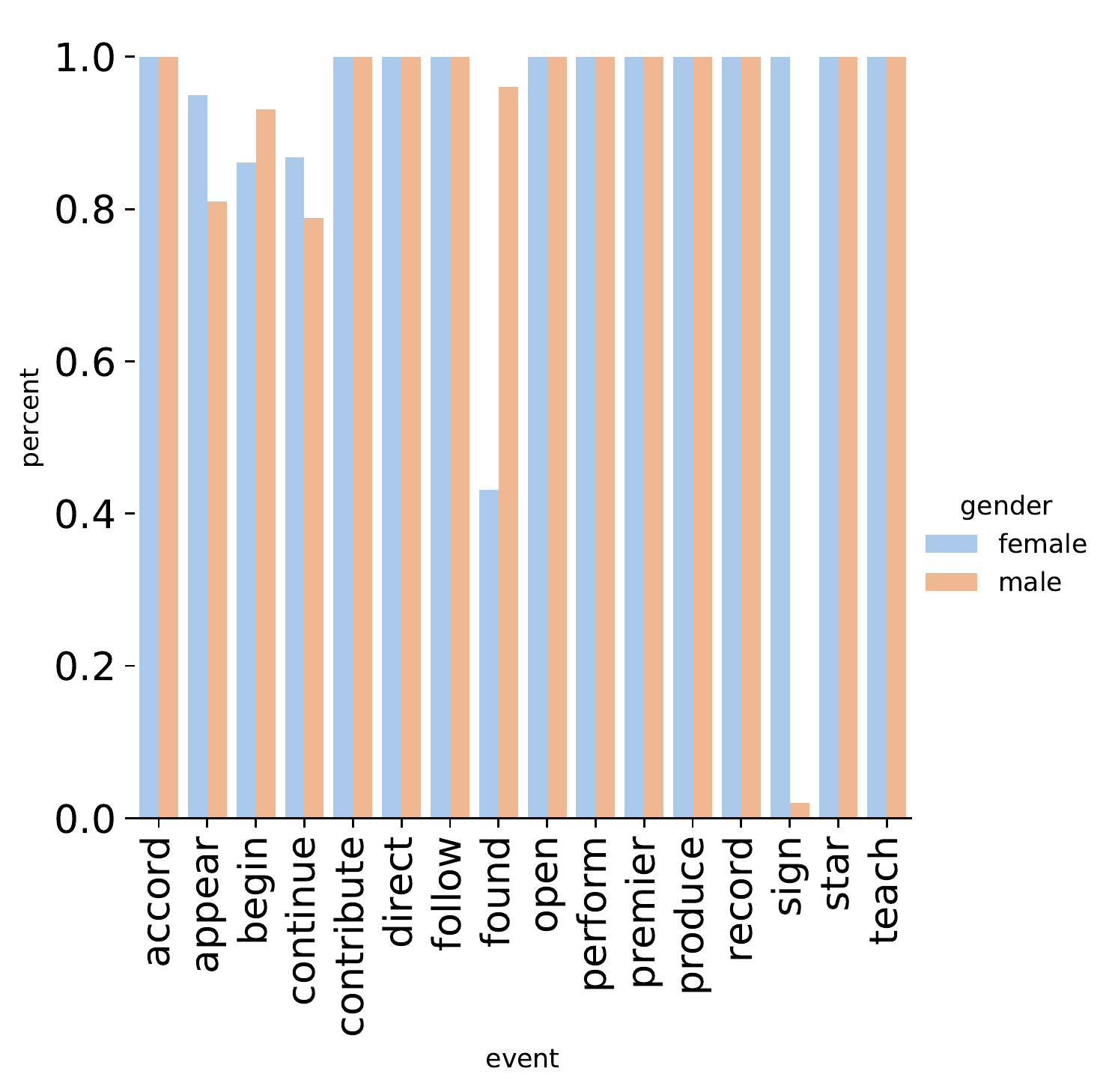}}
\vspace{-0.04cm}
\subfloat[Acting-\emph{c}]{\label{fig:apd_actress} \includegraphics[width=0.23\textwidth, height=0.15\textwidth]{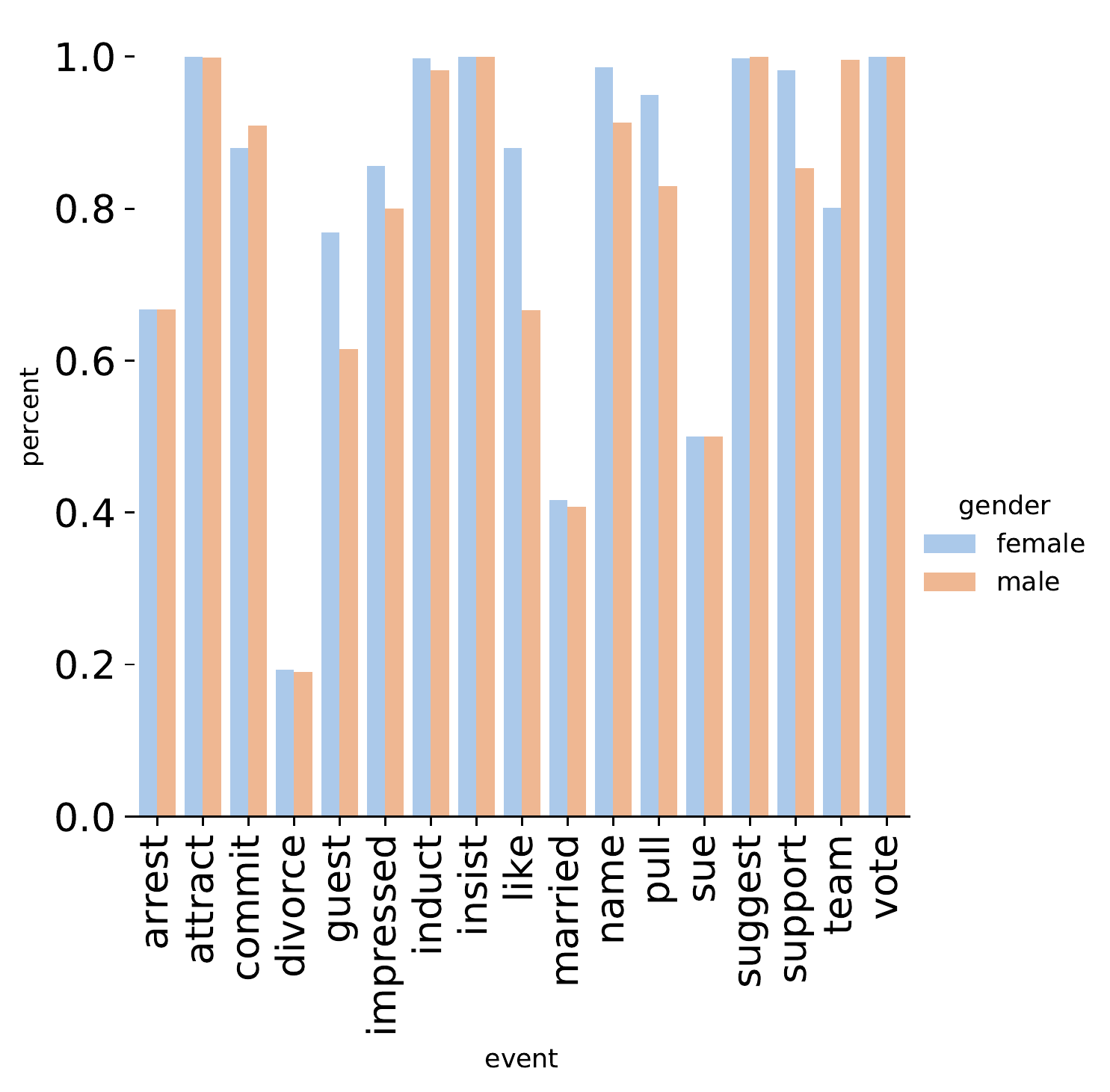}}
\vspace{-0.04cm}
\subfloat[Chefs-\emph{c}]{\label{fig:apd_another} \includegraphics[width=0.23\textwidth, height=0.15\textwidth]{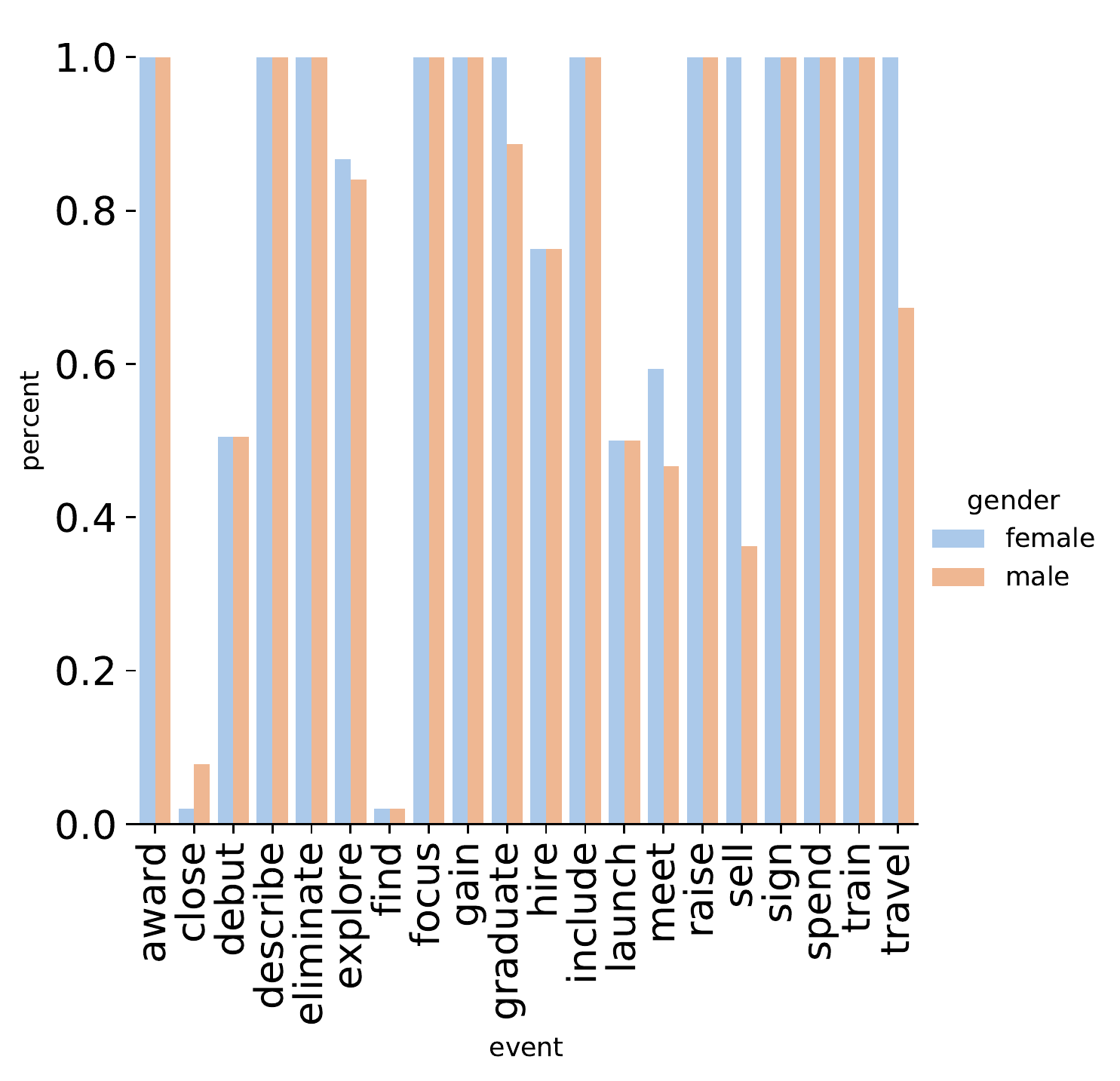}}
\\
\vspace{-0.04cm}
\subfloat[Writers-\emph{pl}]{\label{fig:apd_per_female_writer} \includegraphics[width=0.23\textwidth, height=0.15\textwidth]{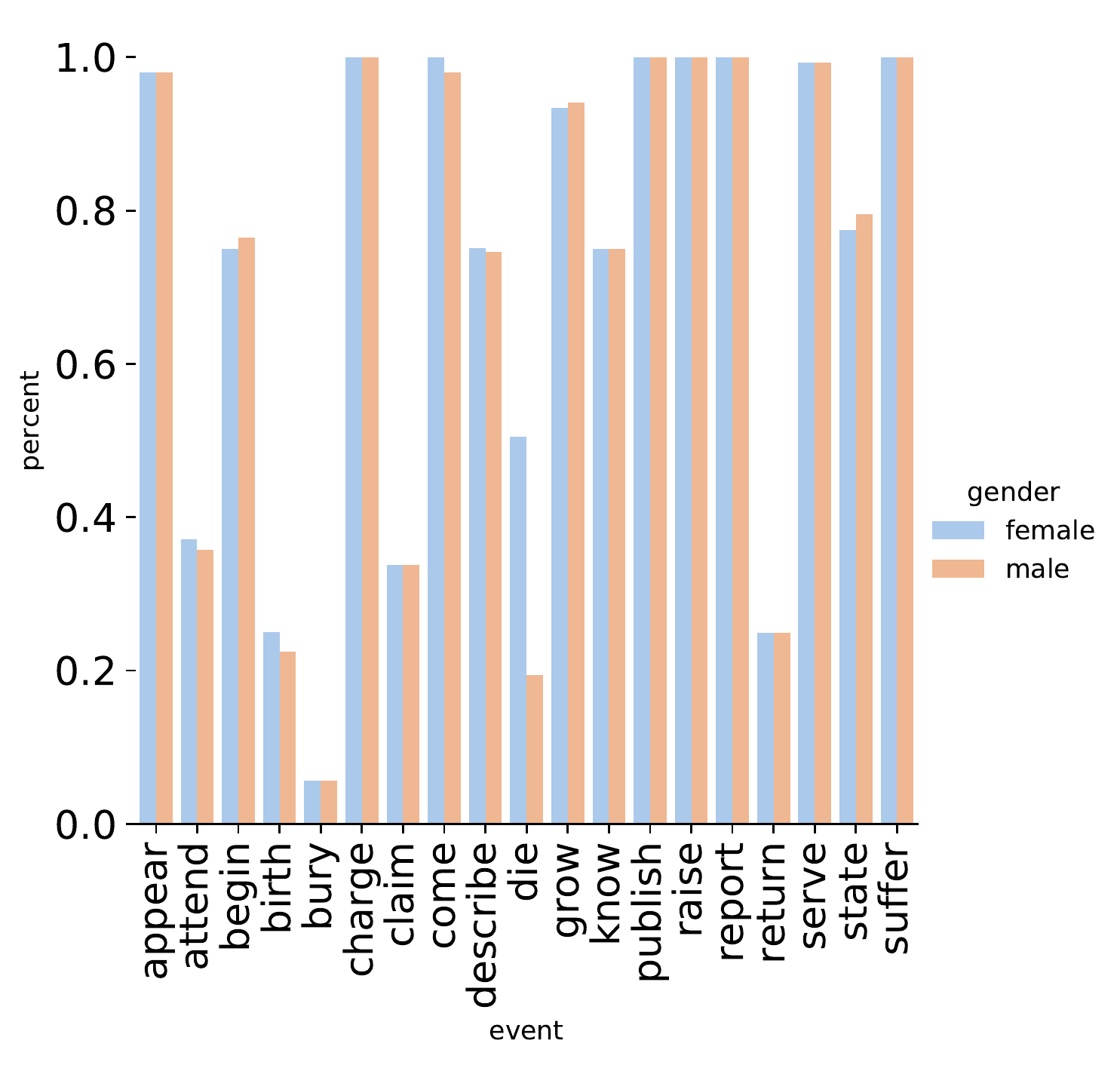}}
\vspace{-0.04cm}
\subfloat[Comedians-\emph{pl}]{\label{fig:apd_per_comedian} \includegraphics[width=0.23\textwidth, height=0.15\textwidth]{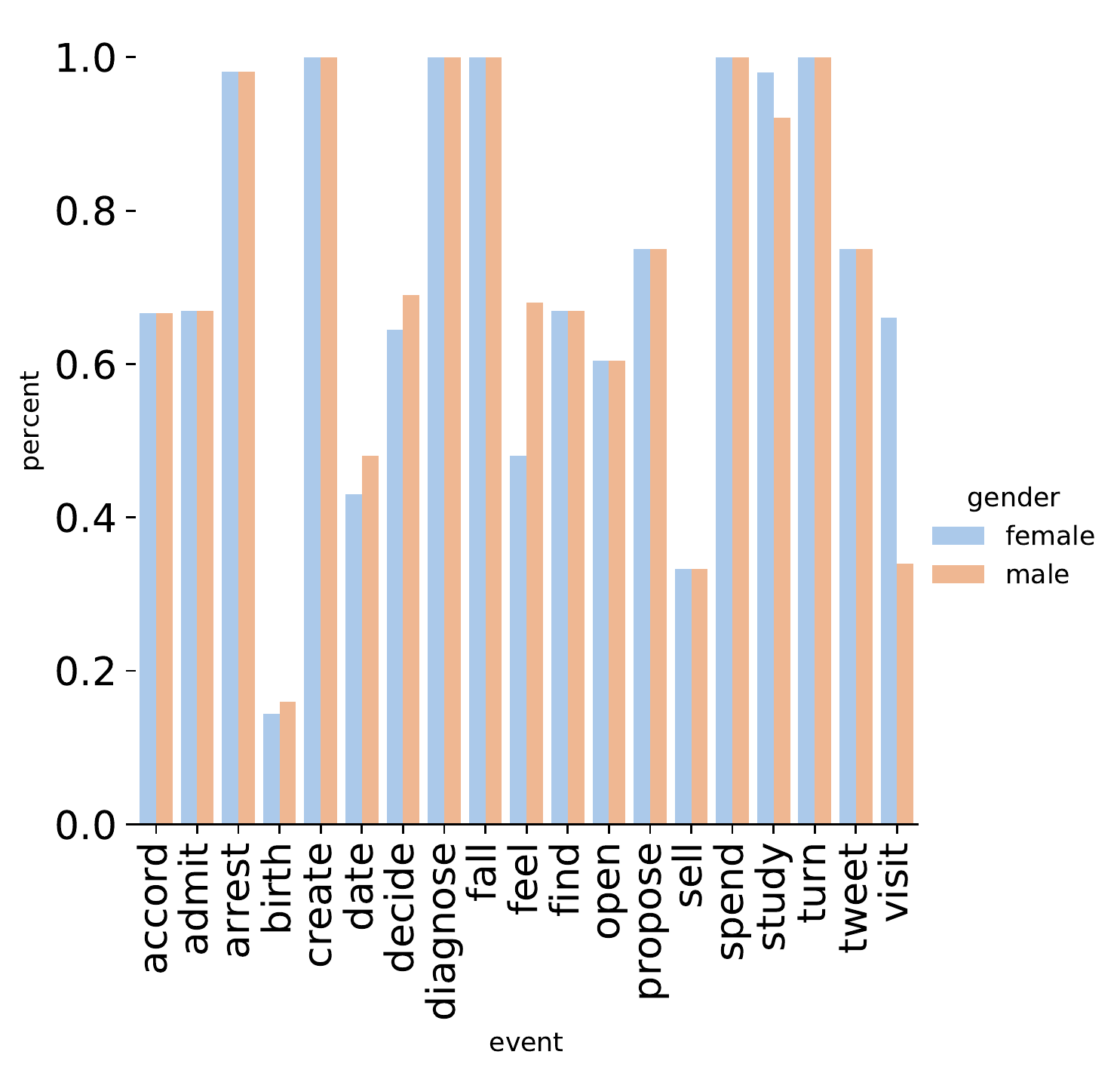}}
\vspace{-0.04cm}
\subfloat[Podcasters-\emph{pl}]{\label{fig:apd_per_podcaster} \includegraphics[width=0.23\textwidth, height=0.15\textwidth]{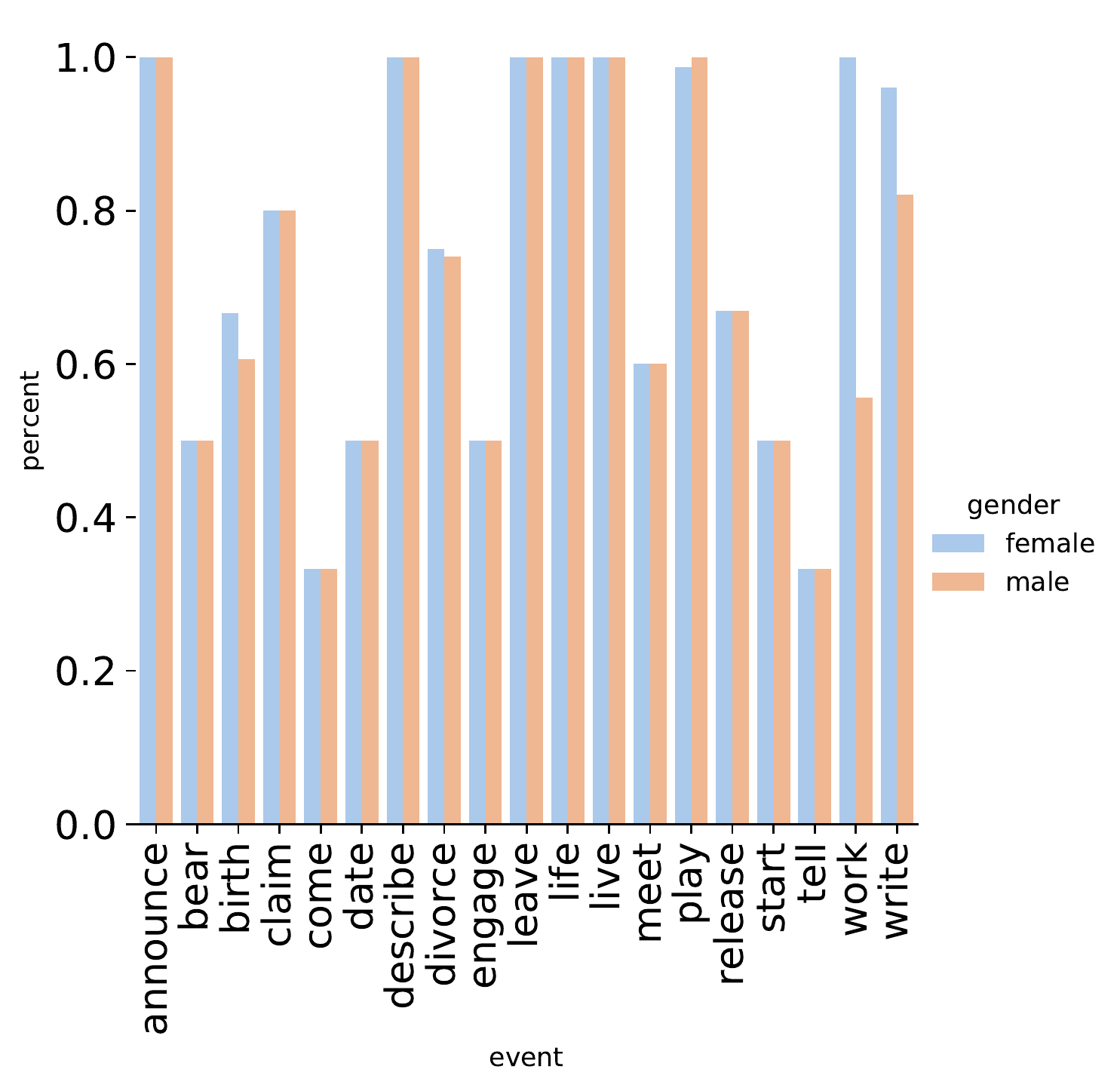}}
\vspace{-0.04cm}
\subfloat[Dancers-\emph{pl}]{\label{fig:apd_per_dancer} \includegraphics[width=0.23\textwidth, height=0.15\textwidth]{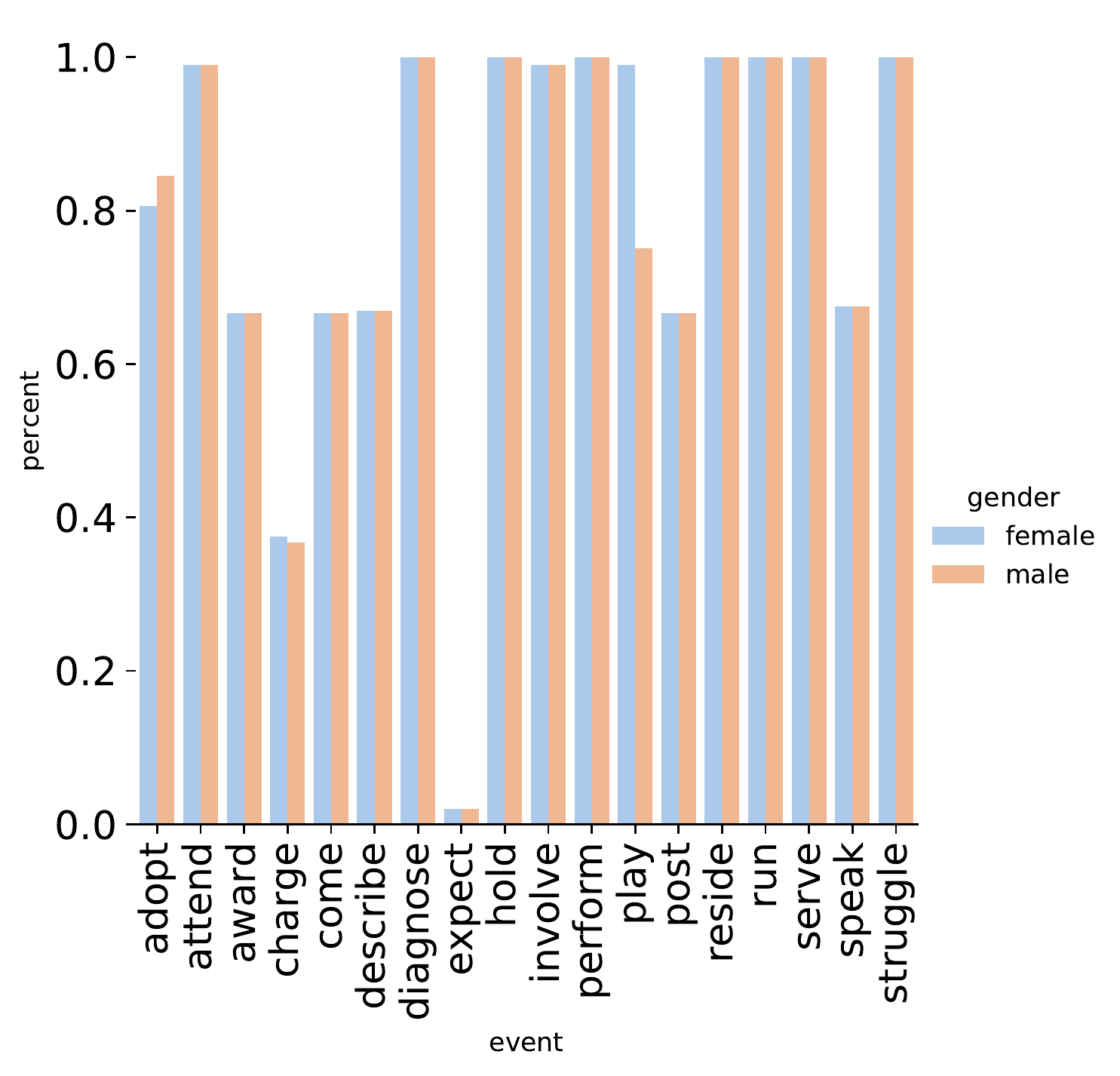}} 
\\
\vspace{-0.04cm}
\subfloat[Chefs-\emph{pl}]{\label{fig:apd_per_female_chef} \includegraphics[width=0.23\textwidth, height=0.15\textwidth]{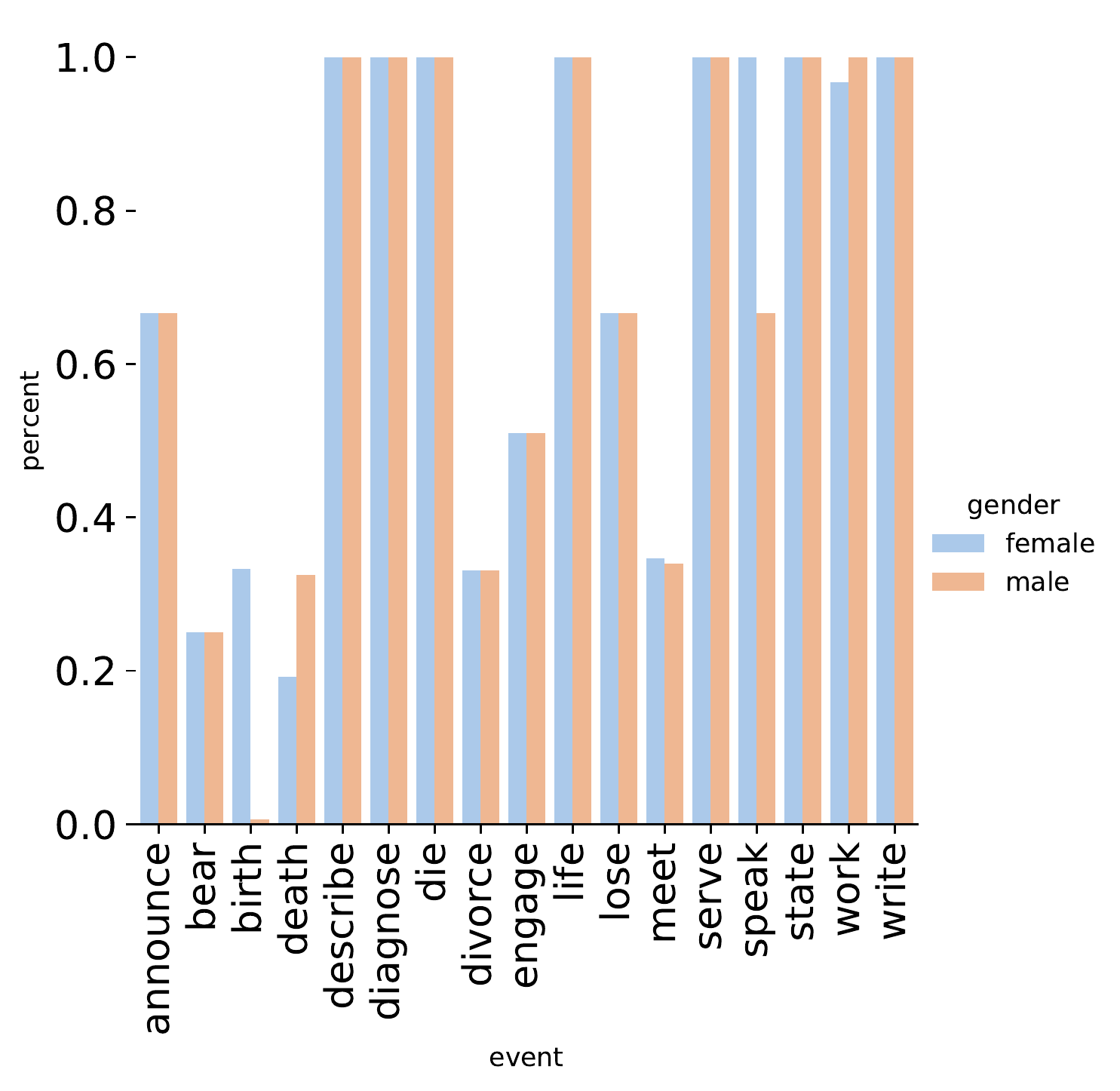}}
\subfloat[Acting-\emph{pl}]{\label{fig:apd_per_actor} \includegraphics[width=0.23\textwidth, height=0.15\textwidth]{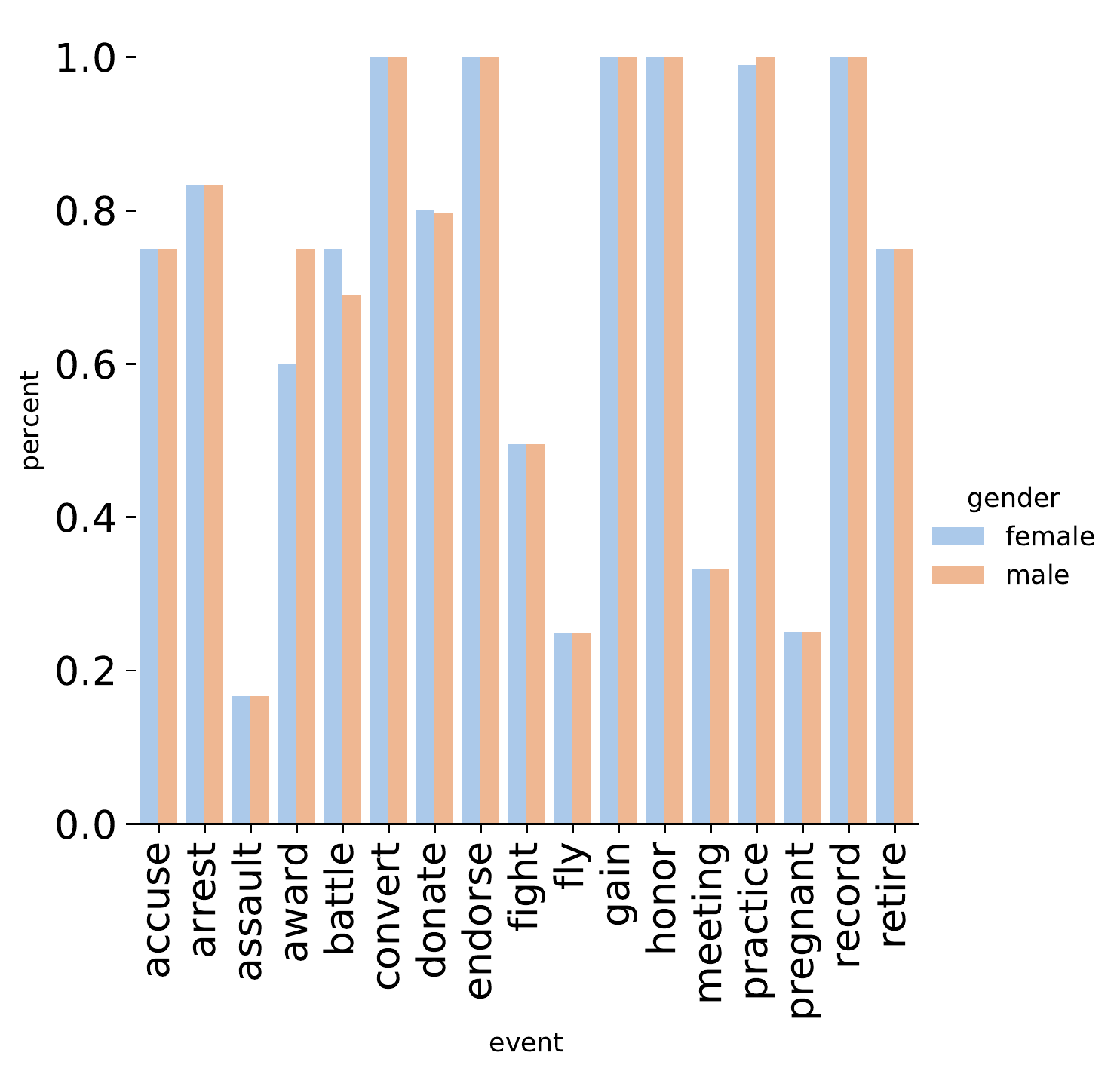}}
\caption{Detection recall on the strategically-generated data. 
(\emph{c}: \emph{Career} section, \emph{pl}: \emph{Personal Life} section)}
\vspace{-1.3cm}
\label{fig:cali}
\end{figure*}

To offset the potential confounding that could be brought by the event extraction model and estimate the actual event frequency of $|e|^{*}$, we use the recall for the event $e$ to calibrate the event frequency $|e|$ for females and males separately. Figure~\ref{fig:cali} shows the calibration result for the 20 most frequent events in our corpus. Please note that Figure~\ref{fig:cali} $(a)$-$(h)$ show the quantitative result for extracted events in the \emph{Career} sections across 8 occupations, and Figure~\ref{fig:cali} $(i)$-$(n)$ for the \emph{Personal Life} sections.

\paragraph{Example Sentence Substitutions for Calibration.} After checking the quality of selected sentences containing the target event trigger, we use 2 steps described in Section~\ref{sec:detect} \emph{Calibration} to compose a synthetic dataset with word substitutions. Here is an example of using \texttt{Name Placeholder}: for target event trigger ``married'' in \texttt{Carole Baskin}'s \emph{Career} section, we have:
\begin{myquote}{0.2in}
    At the age of 17, \underline{Baskin} worked at a Tampa department store. To make money, \underline{she} began breeding show cats; \underline{she} also began rescuing bobcats, and used llamas for a lawn trimming business. In January 1991, \underline{she} married \underline{her} second husband and joined his real estate business.
\end{myquote}

 First, we mark the first name \underline{Baskin} as \texttt{Name Placeholder} and find all gender attributes and pronouns which are consistent with the celebrity's gender.  Then, we replace \underline{Baskin} with 50 female names and 50 male names from \citet{checklist}. If the new name is a male name, we change the corresponding gender attributes (none in this case) and pronouns (e.g., she to he, her to his). 
    
Another example is for the context containing the target event trigger ``married'' in \texttt{Indrani Rahman}'s \emph{Career} section, where there is no first name:
\begin{myquote}{0.2in}
    In 1952, although married, and with a child, \underline{she} became the first \underline{Miss} India, and went on to compete in the \underline{Miss} Universe 1952 Pageant, held at Long Beach, California. Soon, \underline{she} was travelling along with \underline{her} mother and performing all over the world...
\end{myquote}
We replace all pronouns (she to he, her to his) and gender attributes (Miss to Mr).


\begin{table*}[h!]
\small
\centering
\begin{tabular}{l l l l l}
\toprule
Occupation  & Events in Female Career Description & Events in Male Career Description      \\   \midrule

Writer & \begin{tabular}[c]{@{}l@{}} divorce, marriage, involve, organize, wedding, \\ donate, fill, pass, participate, document \end{tabular}   &  \begin{tabular}[c]{@{}l@{}} argue, election, protest, rise, shoot, \\  purchase, kill, host, close, land\end{tabular} \\ \midrule

Acting & \begin{tabular}[c]{@{}l@{}} divorce, wedding, guest, name, commit, \\ attract, suggest, married, impressed, induct \end{tabular}   &  \begin{tabular}[c]{@{}l@{}} support, arrest, war, sue, trial, \\ vote, pull, team, insist, like\end{tabular}  \\\midrule

Comedian & \begin{tabular}[c]{@{}l@{}} birth, eliminate, wedding,  relocate, partner, \\ pursue, impersonate, audition, guest, achieve \end{tabular} &
\begin{tabular}[c]{@{}l@{}} enjoy, hear, cause, buy, conceive, \\ enter, injury, allow, acquire, enter\end{tabular}  \\\midrule

Podcaster & \begin{tabular}[c]{@{}l@{}} land, interview, portray, married, report, \\ earn, praise, talk, shoot, premier \end{tabular} &
\begin{tabular}[c]{@{}l@{}} direct, ask, provide, continue, bring, \\ election, sell, meet, read, open\end{tabular}  \\\midrule

Dancer & \begin{tabular}[c]{@{}l@{}} married, marriage, depart, arrive, organize \\ try, promote, train, divorce, state  \end{tabular} &
\begin{tabular}[c]{@{}l@{}} drop, team, choreograph, explore, break, \\think, add, celebrate, injury, suffer\end{tabular}  \\\midrule

Artist & \begin{tabular}[c]{@{}l@{}}  paint, exhibit, include, return, teach,  \\publish, explore, draw, produce, write \end{tabular} &
\begin{tabular}[c]{@{}l@{}} start, found, feature, award, begin, \\ appear, join, influence, work, create\end{tabular}  \\\midrule

Chef & \begin{tabular}[c]{@{}l@{}} hire, meet, debut, eliminate, sign, \\ graduate, describe, train, begin, appear \end{tabular} &
\begin{tabular}[c]{@{}l@{}} include, focus, explore, award, raise, \\ gain, spend, find, launch, hold\end{tabular}  \\\midrule

Musician & \begin{tabular}[c]{@{}l@{}} run, record, death, found, contribute, \\ continue, perform, teach, appear, accord \end{tabular} &
\begin{tabular}[c]{@{}l@{}} sign, direct, produce, premier, open, \\  announce, follow, star, act, write\end{tabular}  \\\midrule

\end{tabular}
\caption{The top 10 extracted events in \emph{Career} section. }
\label{tab:add}
\end{table*}


\begin{table*}[h!]
\small
\centering
\begin{tabular}{l l l l l}
\toprule
Occupation  & Events in Female Personal Life Description & Events in Male Personal Life Description      \\   \midrule

Writer & \begin{tabular}[c]{@{}l@{}} bury, birth, attend, war, grow, \\ serve, appear, raise, begin, divorce \end{tabular}   &  \begin{tabular}[c]{@{}l@{}} know, report, come, charge, publish, \\ claim, suffer, return, state, describe\end{tabular} \\ \midrule

Acting &   \begin{tabular}[c]{@{}l@{}} pregnant, practice, wedding, record, \\ convert,  honor, gain, retire, rap, bring\end{tabular} & \begin{tabular}[c]{@{}l@{}} accuse, trip, fly, assault, endorse, \\ meeting, donate, fight, arrest, found \end{tabular}     \\\midrule

Comedian & \begin{tabular}[c]{@{}l@{}} feel, birth, fall, open, decide, \\ date, diagnose, tweet, study, turn \end{tabular} &
\begin{tabular}[c]{@{}l@{}} visit, create, spend, propose, lawsuit, \\ accord, arrest, find, sell, admit\end{tabular}  \\\midrule

Podcaster & \begin{tabular}[c]{@{}l@{}} date, describe, tell, life, come, \\ leave, engage, live, start, reside \end{tabular} &
\begin{tabular}[c]{@{}l@{}} play, write, bear, release, claim, \\ birth, divorce, meet, announce, work\end{tabular}  \\\midrule

Dancer & \begin{tabular}[c]{@{}l@{}} marry, describe, diagnose, expect, speak, \\ post, attend, come, play, reside  \end{tabular} &
\begin{tabular}[c]{@{}l@{}} hold, involve, award, run, serve, \\ adopt, charge, suit, struggle, perform\end{tabular}  \\\midrule

Chef & \begin{tabular}[c]{@{}l@{}}death, serve, announce, describe, born, \\ die, life, state, marriage, live \end{tabular} &
\begin{tabular}[c]{@{}l@{}} birth, lose, divorce, speak, meet, \\ work, diagnose, wedding, write, engage\end{tabular}  \\\midrule
\end{tabular}
\caption{The top 10 extracted events in \emph{Personal Life} section. }
\label{tab:add_pl}
\end{table*}

\paragraph{Interpret the Quantitative Calibration Result.} We use the calibration technique to calibrate potential gender biases from the model that could have complicated the analysis. In Figure~\ref{fig:cali}, we can see that there is little gender bias at the model level: the model has the same performance for females and males among most events. 

Besides, we notice that the model fails to detect and has a low recall for few events in the generated synthetic dataset. We speculate that this is because of the brittleness in event extraction models triggered by the word substitution. We will leave more fine-grained analysis at the model level for future work.
We focus on events for which the model performs largely differently for females and males during our calibration. Thus, we select and focus on the events that have different performances for females and males over a threshold, which we take 0.05 during our experiment, to calibrate the analysis result.
\subsection{Top Ten Extracted Events}
\label{appendix:events}

Table~\ref{tab:add} and Table~\ref{tab:add_pl} show the top 10 events and serves as the supplement of top 5 events that we reported for \emph{Career} and \emph{Personal Life} sections.  

\subsection{Details for Calculating WEAT Score}
\label{appendix:weat}

The WEAT score is in the range of $-2$ to $2$. A high positive score indicates that extracted events for females are more associated with female attributes in the embedding space. A high negative score means that extracted events for females are more associated with male attributes. To calculate the WEAT score, we input two lists of extracted events for females $E_f$, and males $E_m$, together with two lists of gender attributes $A$ and $B$, then calculate:
\begin{equation}
\small
    S(E_f,E_m,A,B) = \sum_{e_f\in E_f} s(e_f, A, B) - \sum_{e_m\in E_m} s(e_m, A, B),
\end{equation}
where 
\begin{equation}
\small
   s(w, A, B)=\text{mean}_{a\in A} cos(\vec{w}, \vec{a})-\text{mean}_{b\in B}cos(\vec{w}, \vec{b}).
\end{equation} 
Following \citet{WEAT}, we have ``female, woman, girl, sister, she, her, hers, daughter'' as female attribute list $A$ and ``male, man, boy, brother, he, him, his, son'' as male attributes list $B$. To calculate WEAT$^{*}$, we replace the input lists $E_f$ and $E_m$ with all non-stop words tokens in raw texts from either \emph{Career} section or \emph{Personal Life} section. 

\subsection{Extracted Events Frequency Distribution}
\label{appendix:distribution}
\begin{figure*}[h]
  \vspace{-1cm}
        \centering
        \vspace{-0.2cm}
        \makebox[\textwidth][c]{\parbox{0.98\textwidth}{
        \subfloat[Male Comedians-\emph{c}]  {
                \includegraphics[width=0.23\textwidth, height=0.15\textwidth]{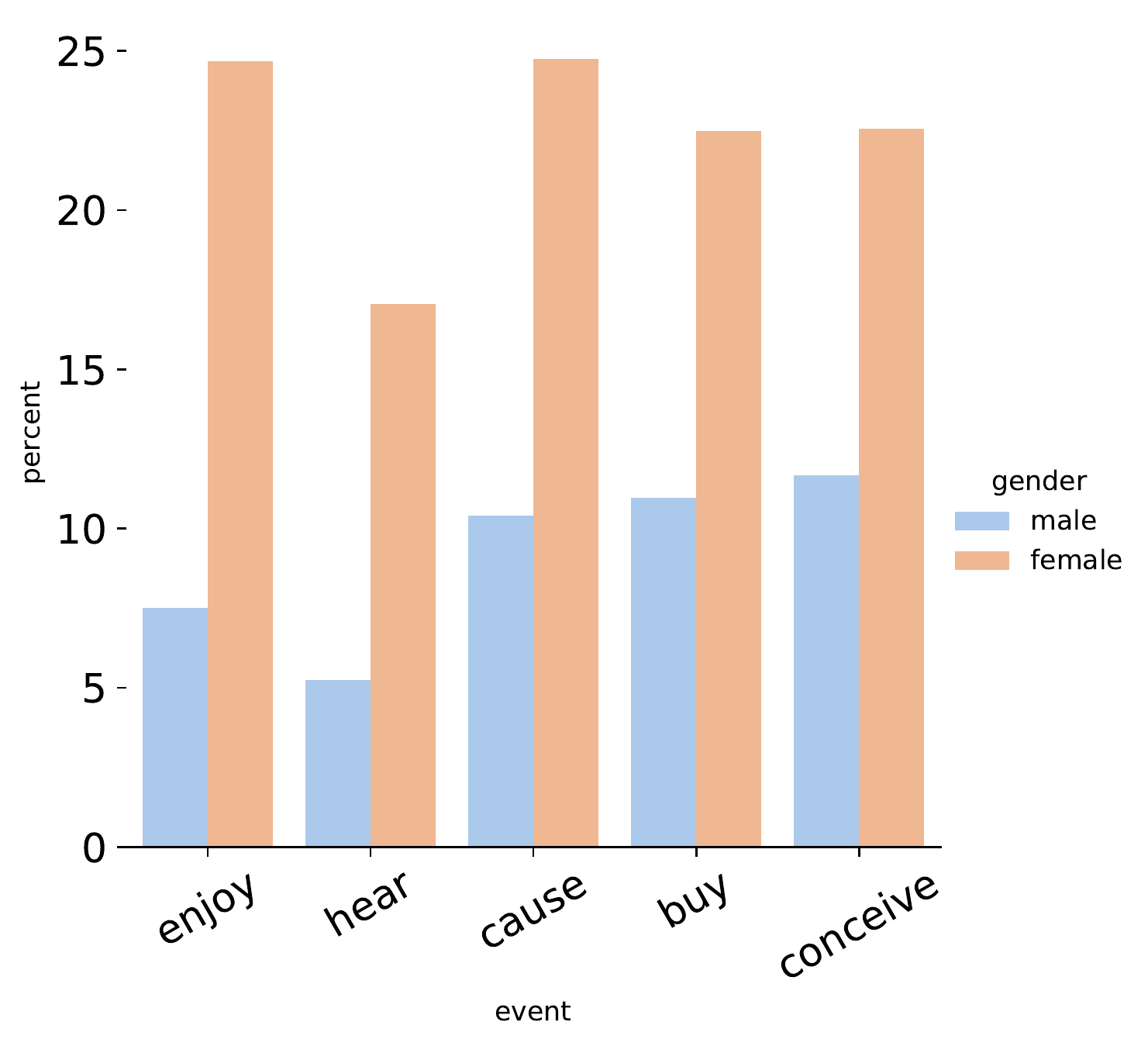}

        }
        \subfloat[Female Comedians-\emph{c}] {
                \includegraphics[width=0.23\textwidth, height=0.15\textwidth]{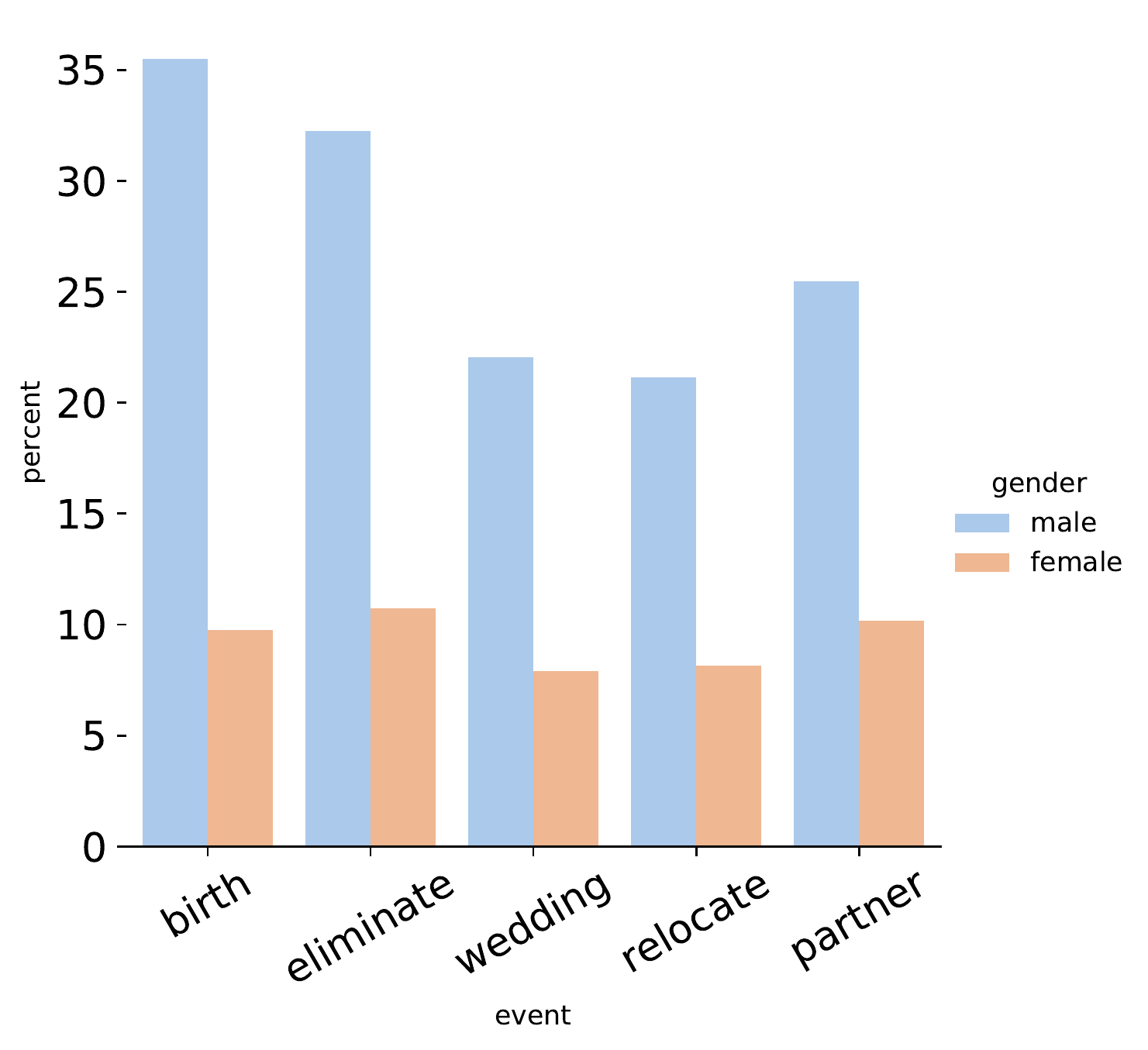}
            }
        \subfloat[Male Dancers-\emph{c}]  {
                \includegraphics[width=0.23\textwidth, height=0.15\textwidth]{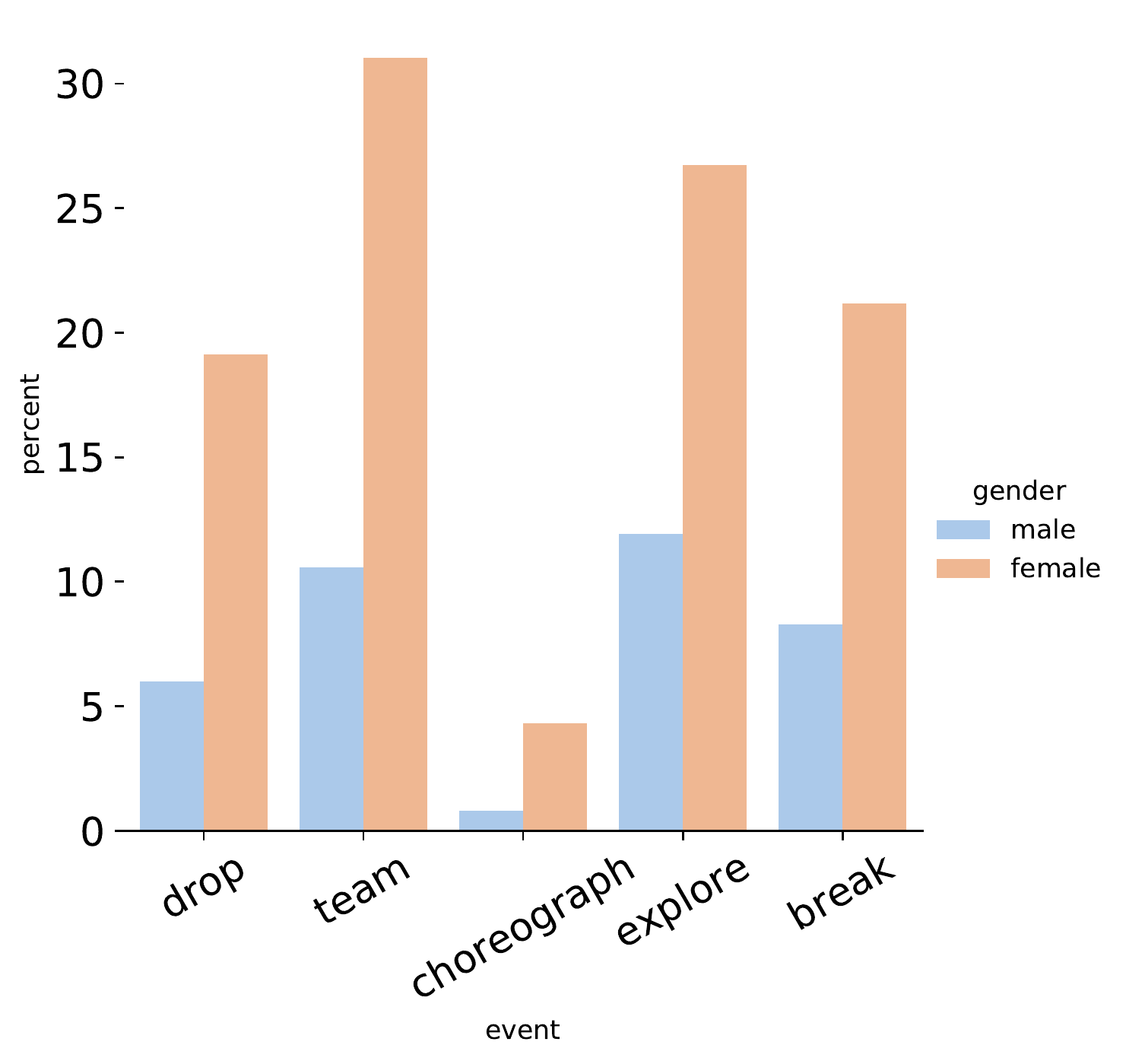}
        }
        \subfloat[Female Dancers-\emph{c}] {
                \includegraphics[width=0.23\textwidth, height=0.15\textwidth]{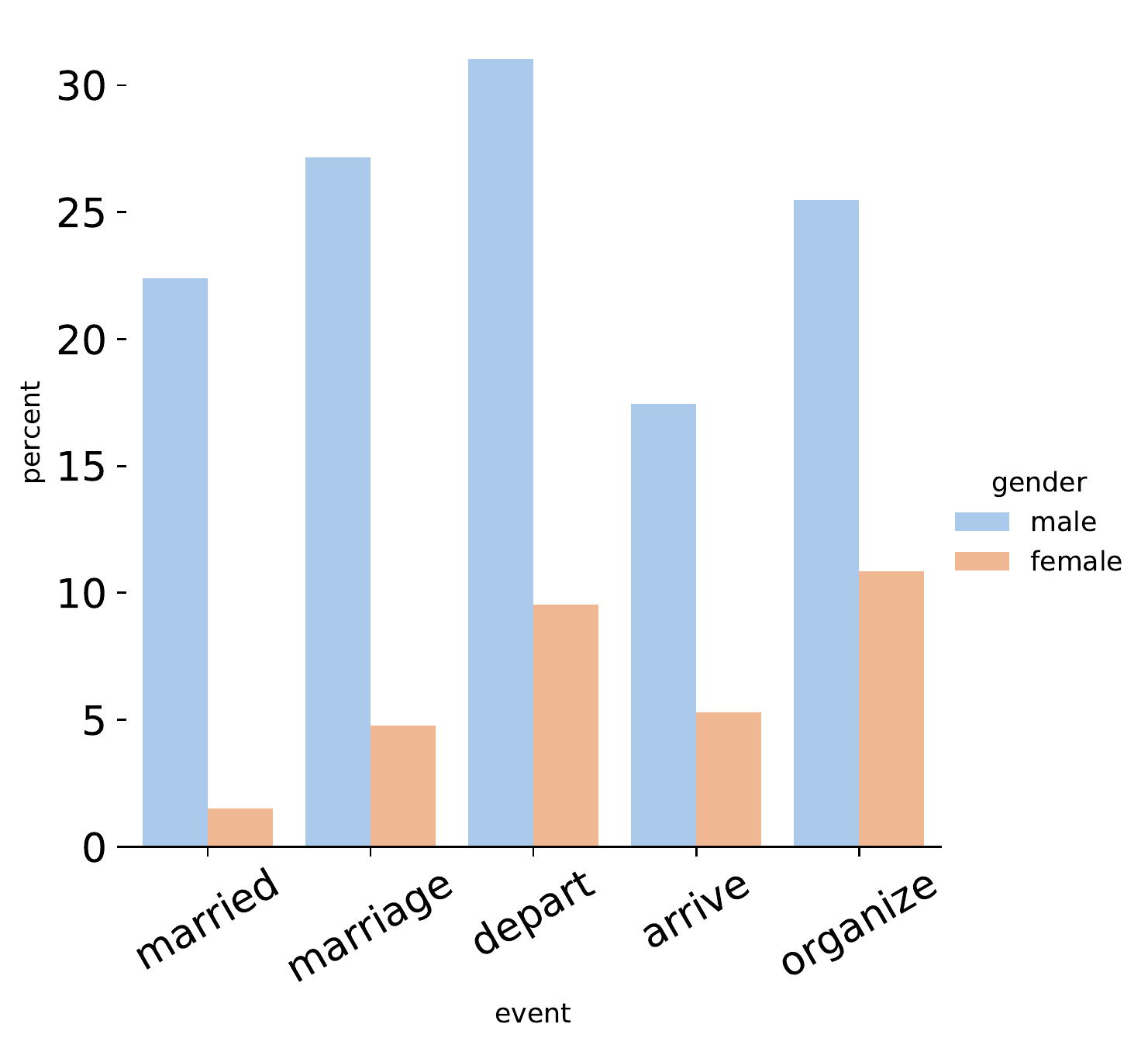}
        }
        \\
        \vspace{-0.2cm}
        \subfloat[Male Podcasters-\emph{c}] {
            \includegraphics[width=0.23\textwidth, height=0.15\textwidth]{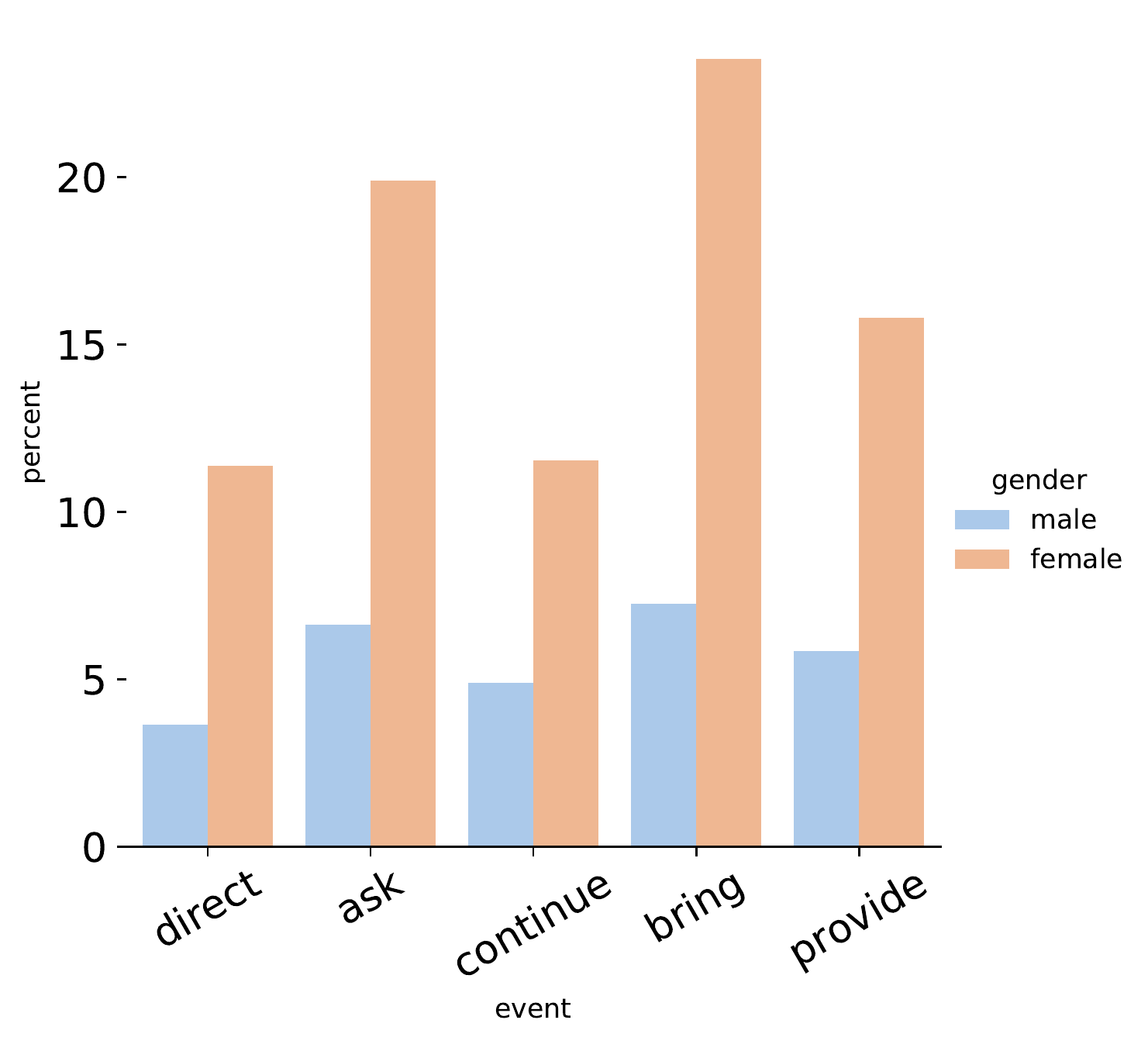}
        }
        \subfloat[Female Podcasters-\emph{c}] {
                \includegraphics[width=0.23\textwidth, height=0.15\textwidth]{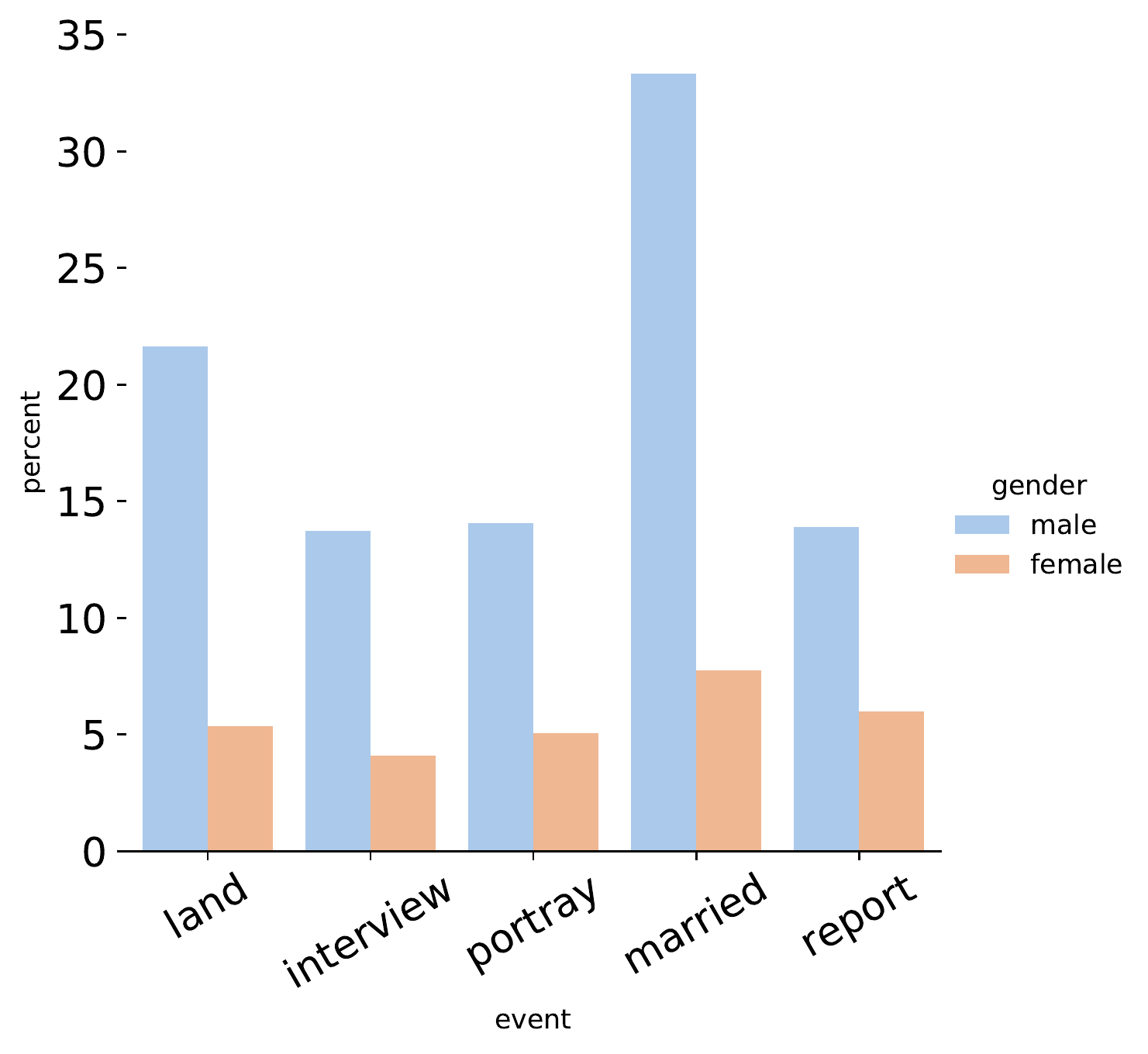}
            }   
        \subfloat[Male Chefs-\emph{c}]  {
                \includegraphics[width=0.23\textwidth, height=0.15\textwidth]{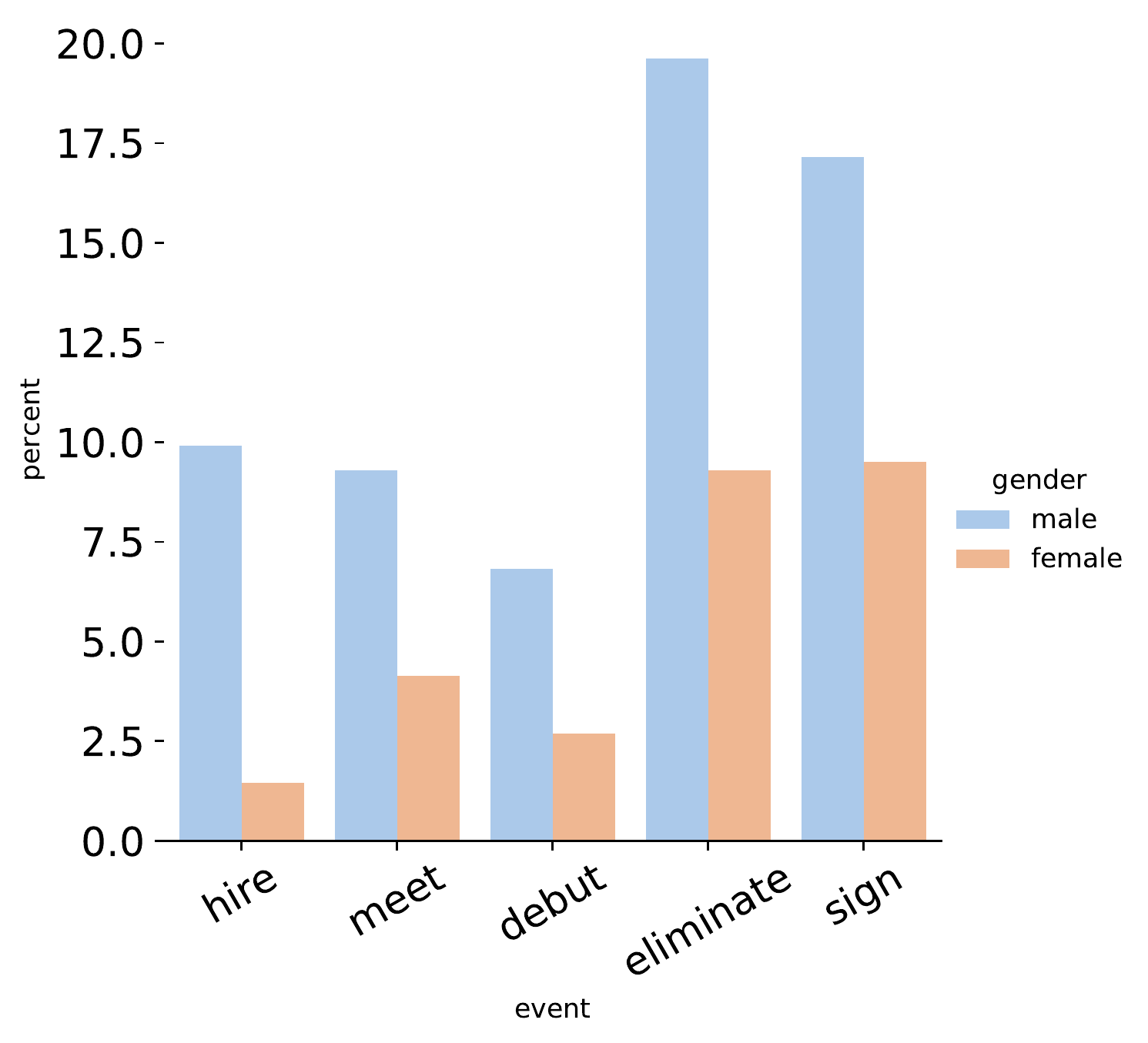}

        }
        \subfloat[Female Chefs-\emph{c}] {
                \includegraphics[width=0.23\textwidth, height=0.15\textwidth]{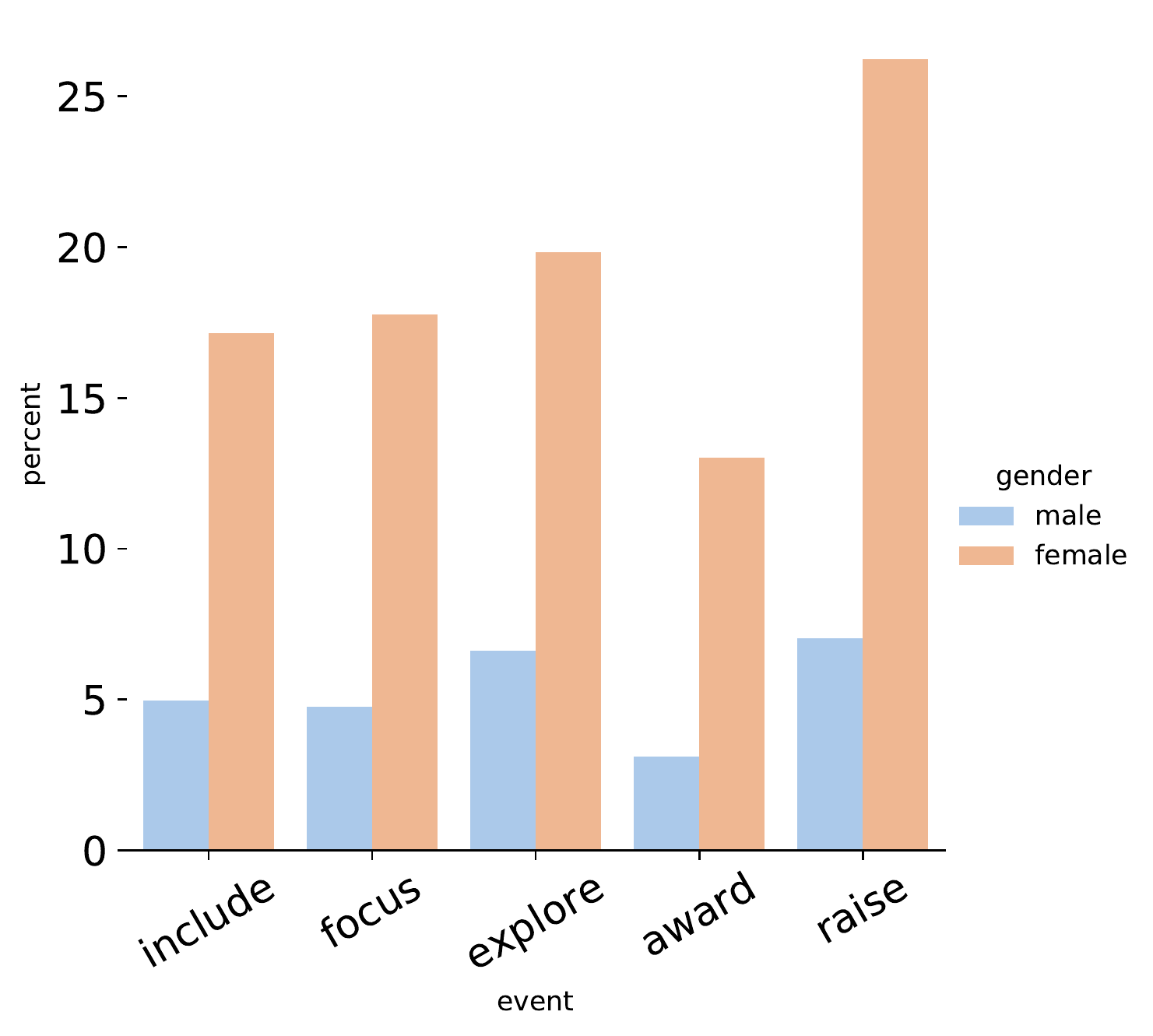}
        }
        \\
        \vspace{-0.2cm}
        \subfloat[Male Artists-\emph{c}]  {
                \includegraphics[width=0.23\textwidth, height=0.15\textwidth]{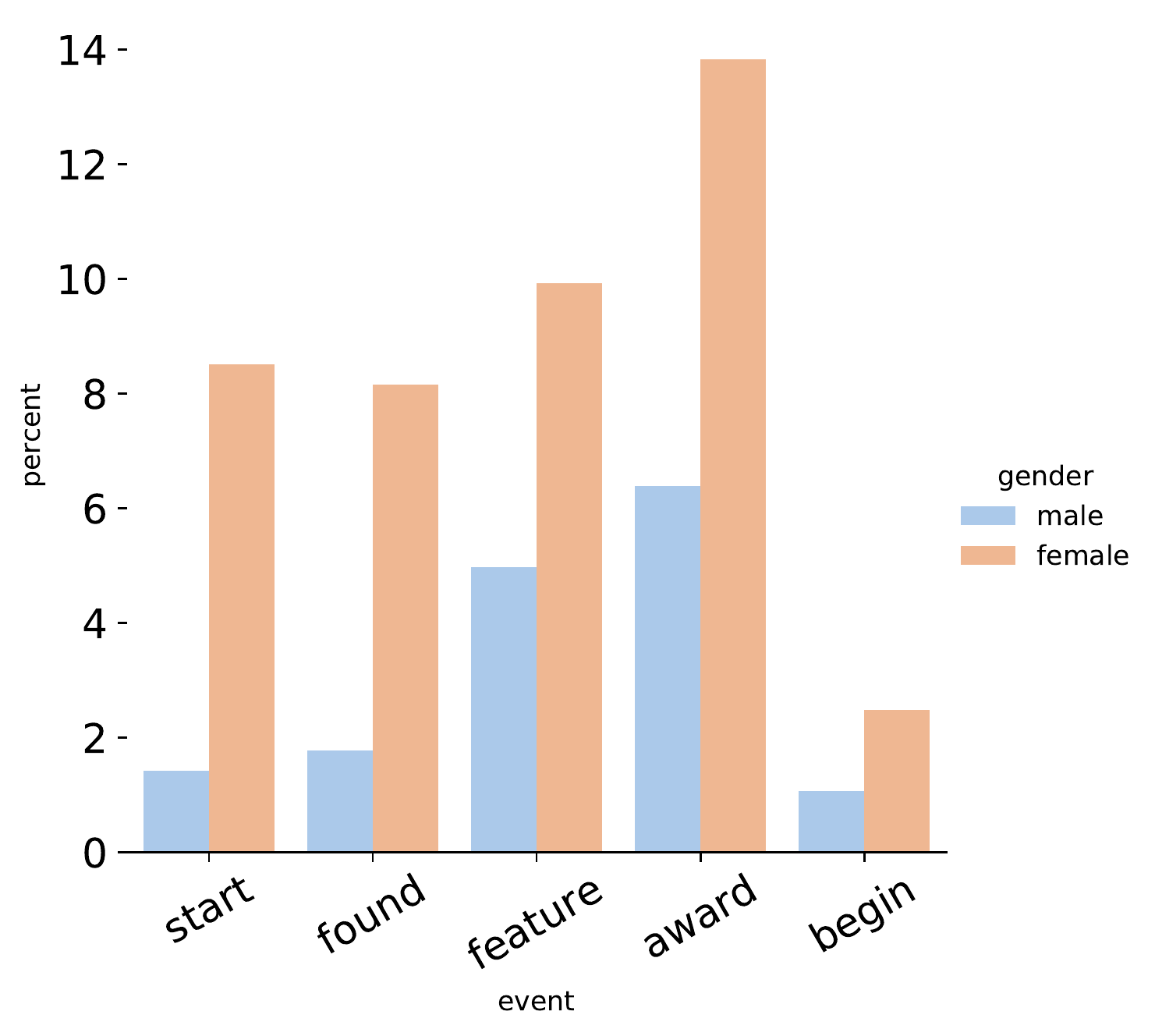}
        }
        \subfloat[Female Artists-\emph{c}] {
                \includegraphics[width=0.23\textwidth, height=0.15\textwidth]{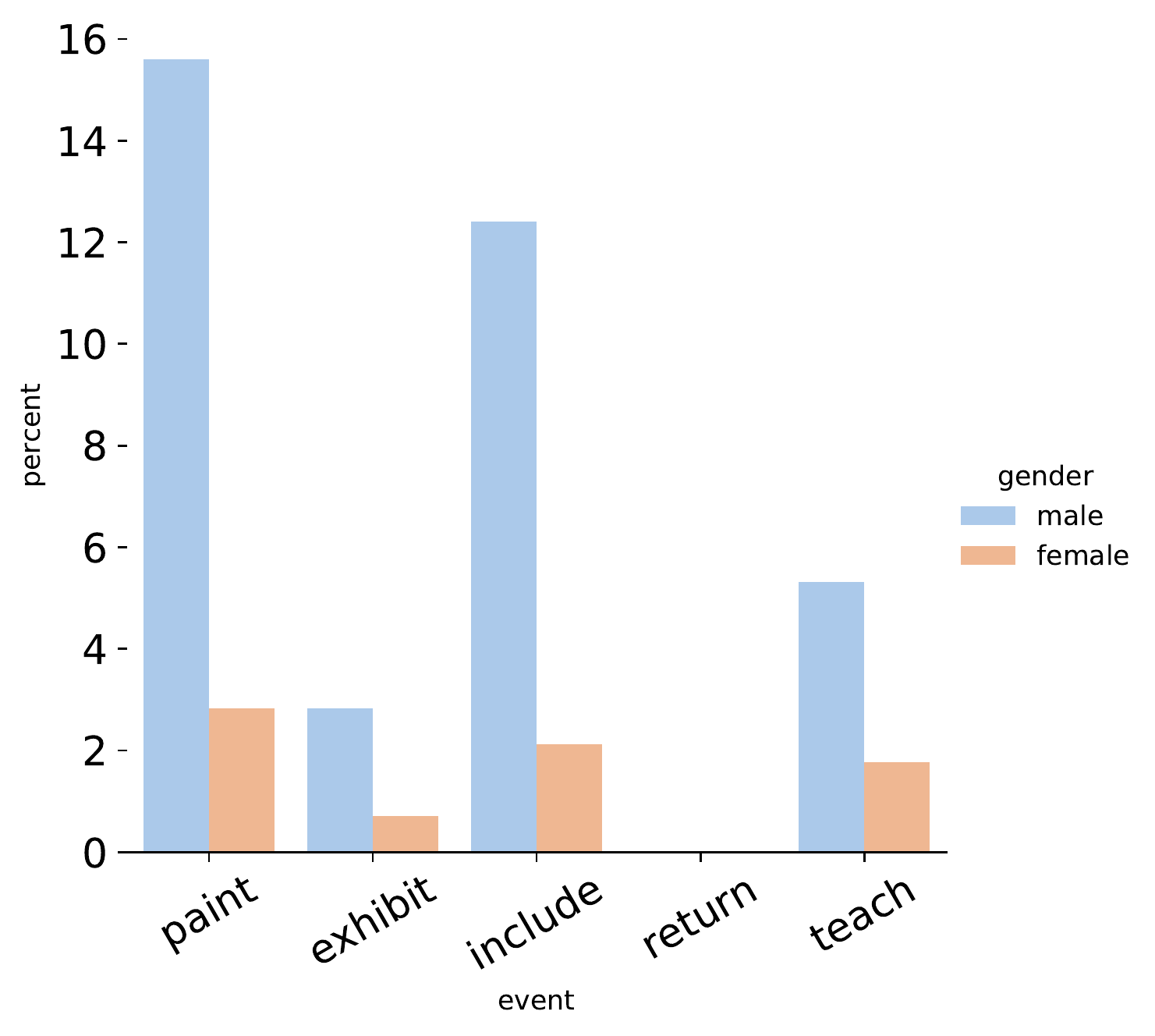}
            }
        \subfloat[Male Musicians-\emph{c}] {
            \includegraphics[width=0.23\textwidth, height=0.15\textwidth]{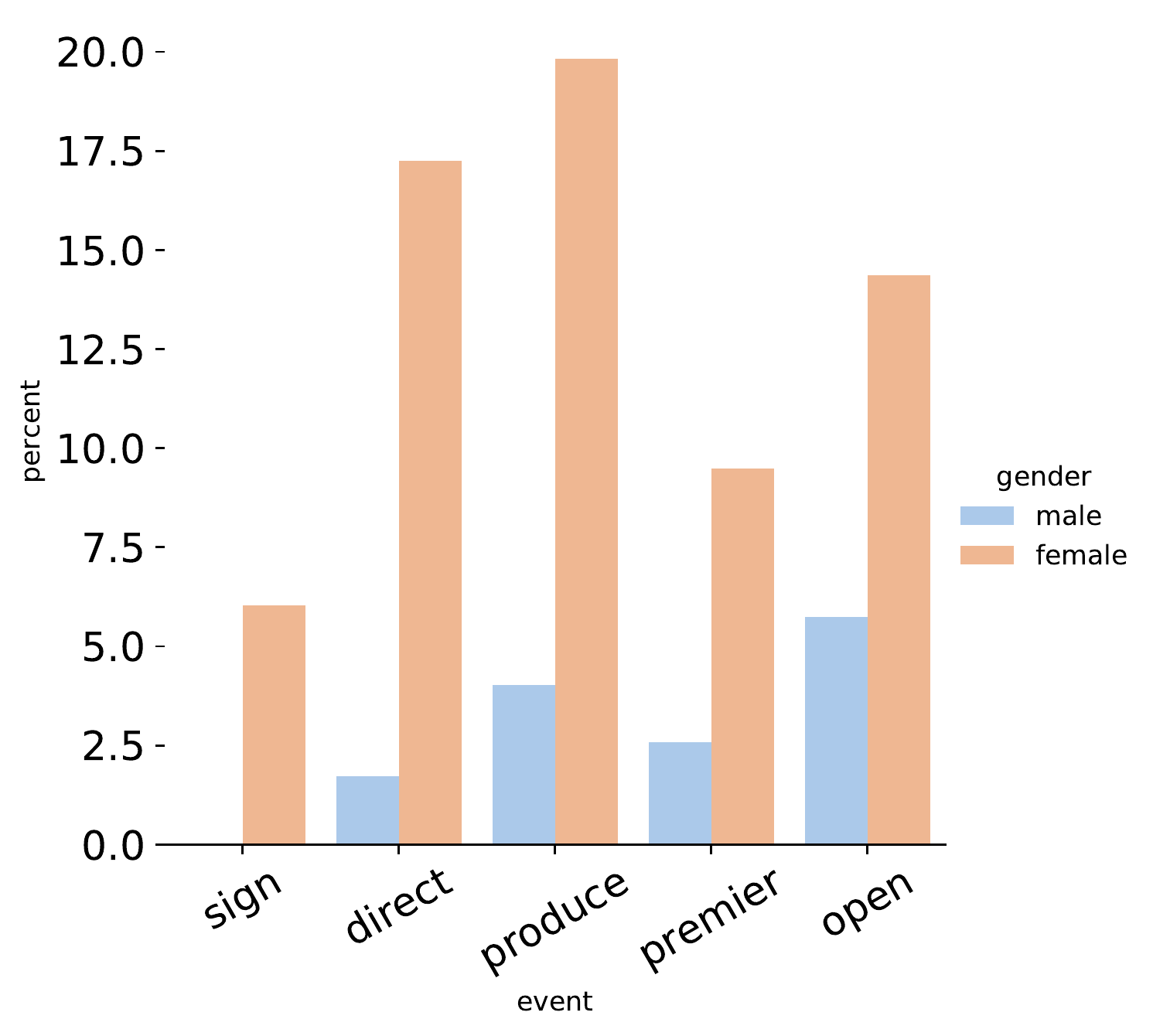}
        }
        \subfloat[Female Musicians-\emph{c}] {
             \includegraphics[width=0.23\textwidth, height=0.15\textwidth]{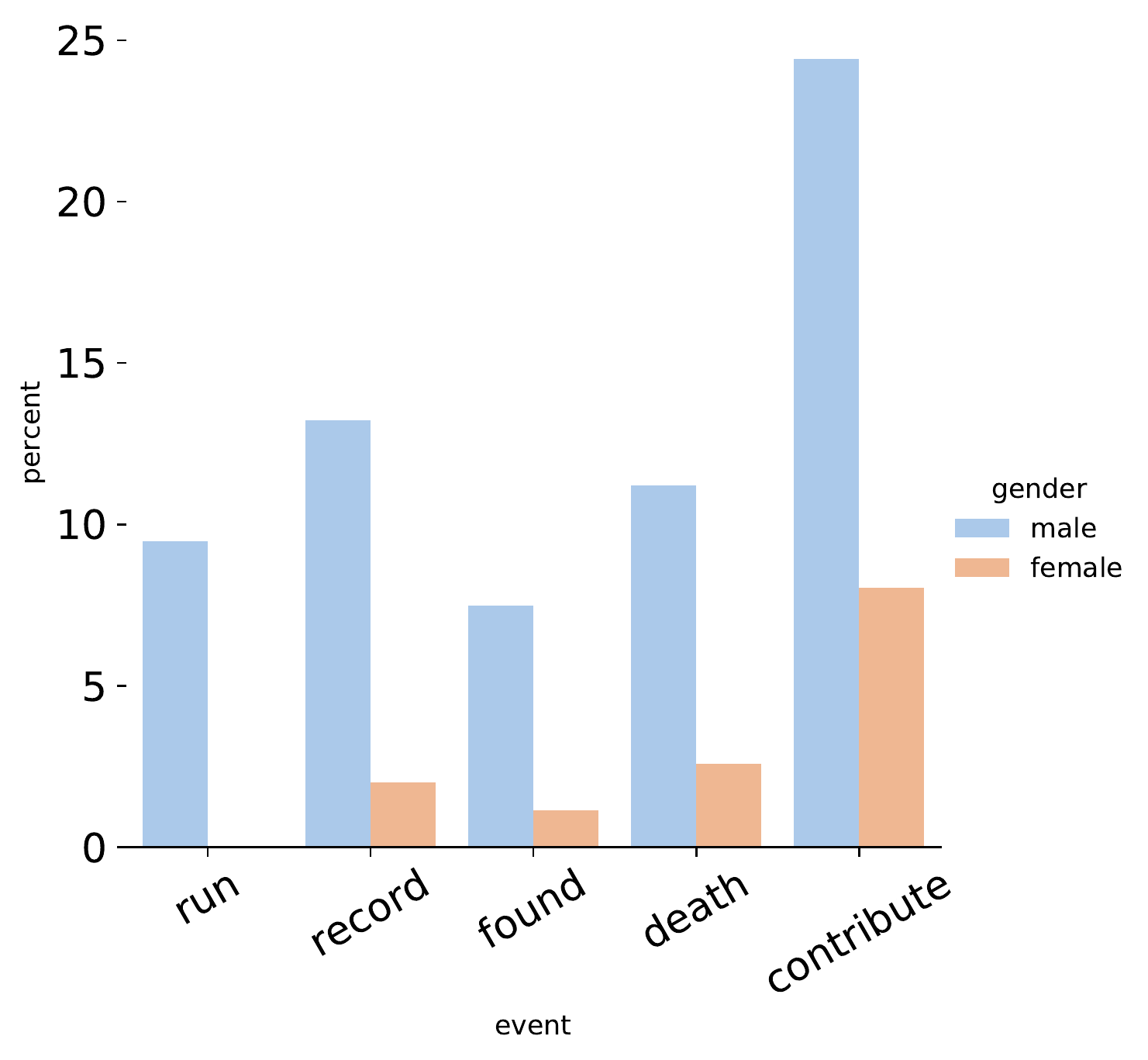}
            }  
        \\
        \vspace{-0.2cm}
        
        \subfloat[Male Comedians-\emph{pl}]  {
                \includegraphics[width=0.23\textwidth, height=0.15\textwidth]{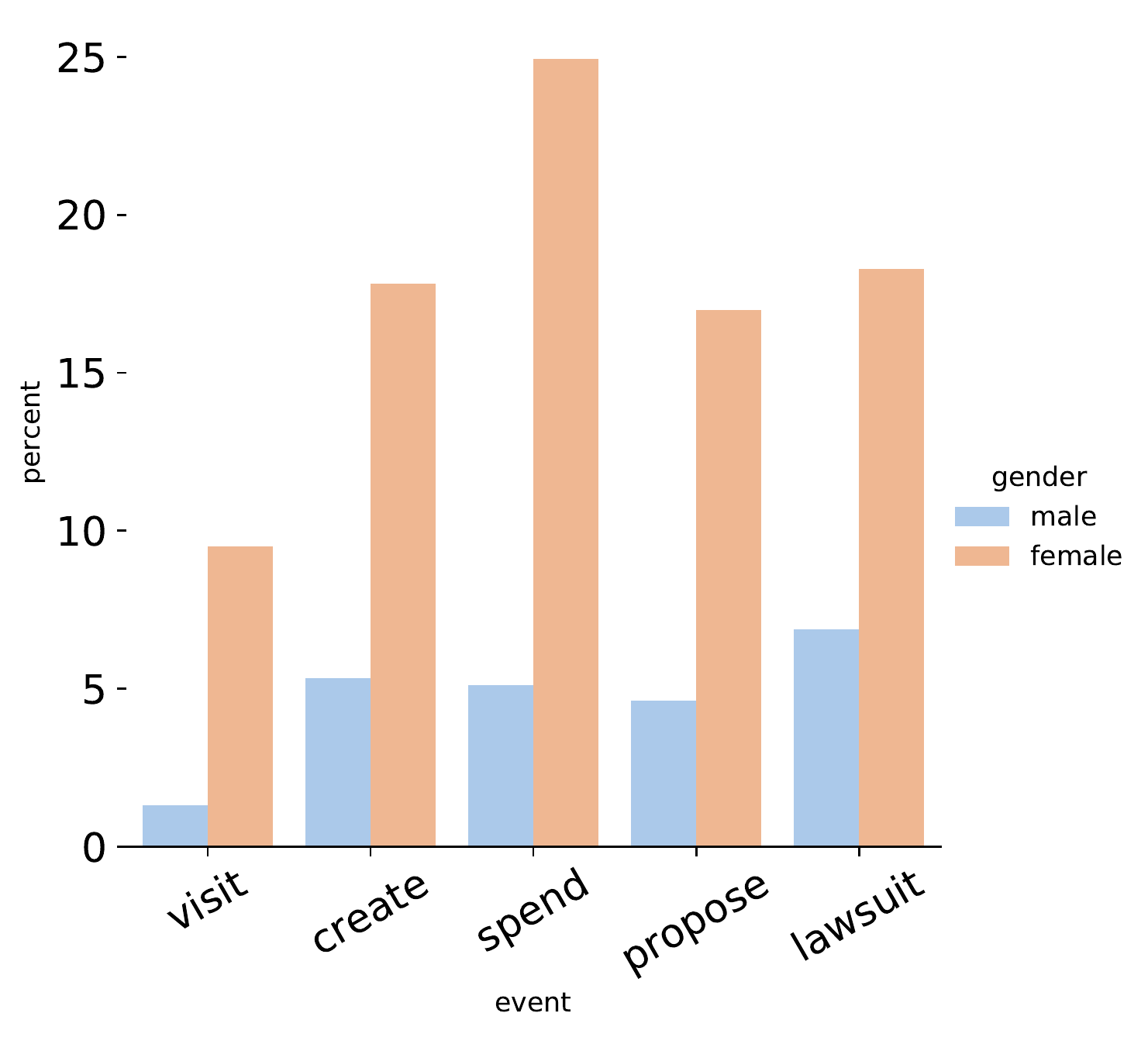}

        }
        \subfloat[Female Comedians-\emph{pl}] {
                \includegraphics[width=0.23\textwidth, height=0.15\textwidth]{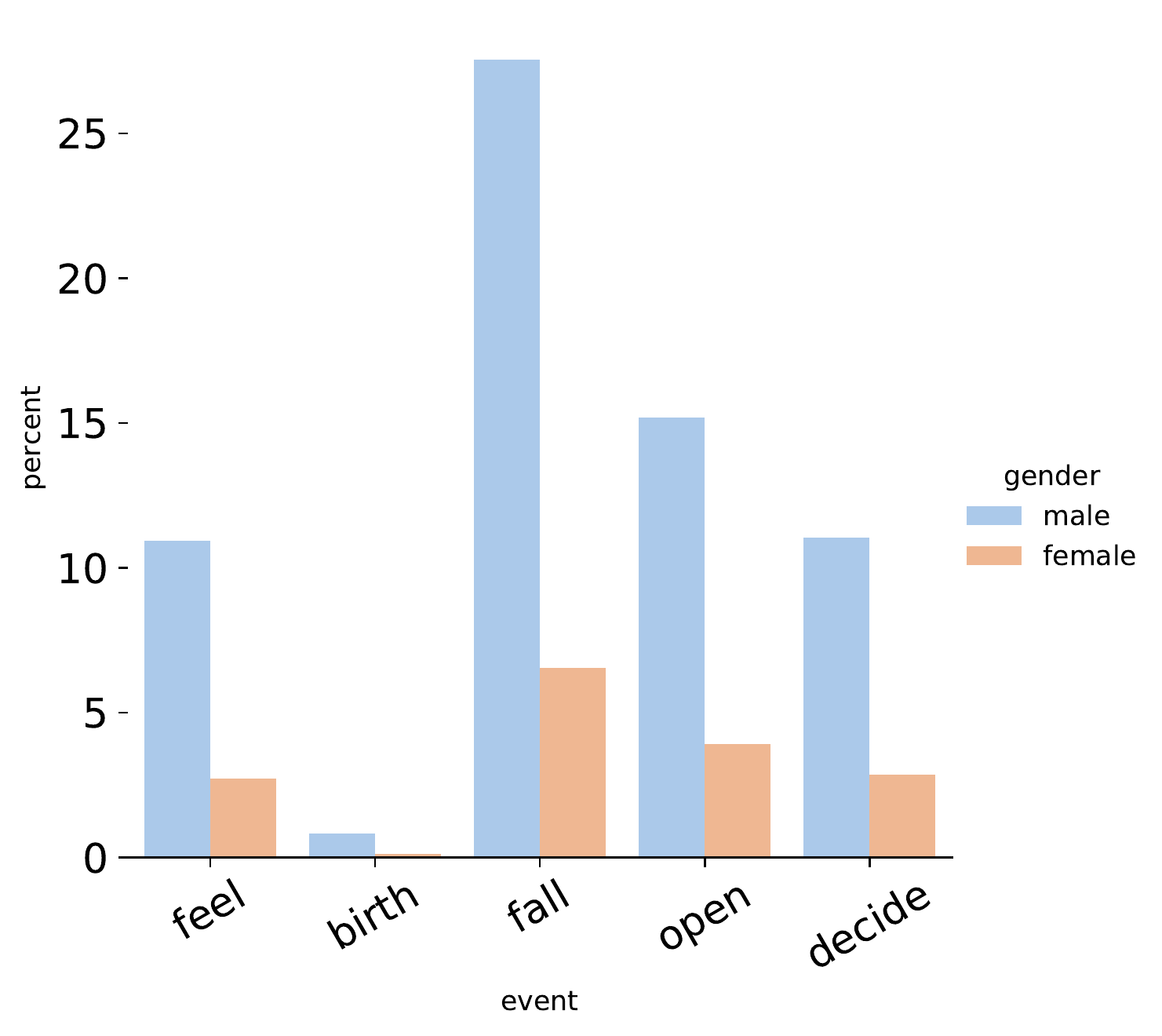}
            }
        \subfloat[Male Dancers-\emph{pl}]  {
                \includegraphics[width=0.23\textwidth, height=0.15\textwidth]{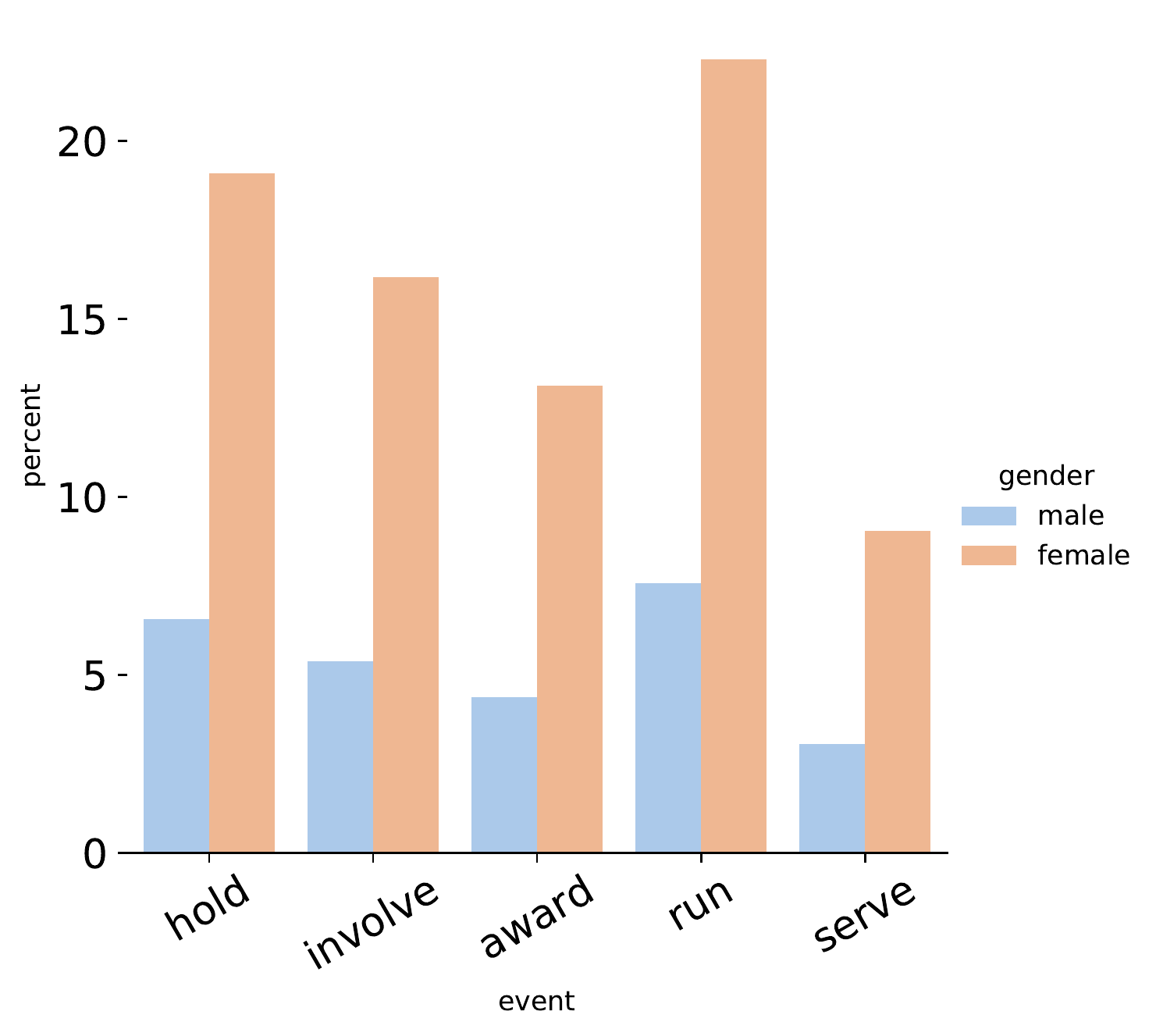}
        }
        \subfloat[Female Dancers-\emph{pl}] {
                \includegraphics[width=0.23\textwidth, height=0.15\textwidth]{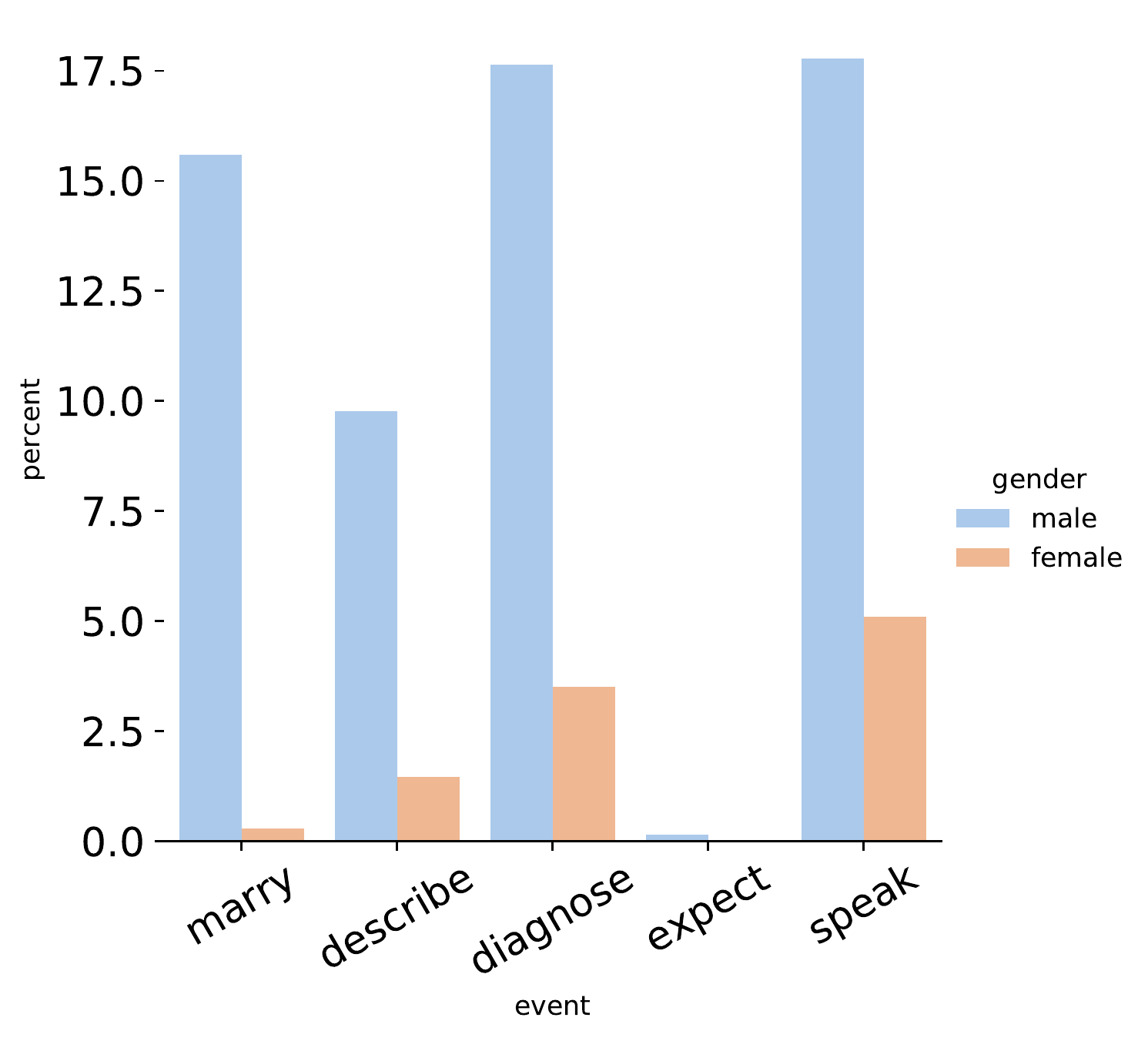}
            }
            \\
        \vspace{-0.2cm}
        \subfloat[Male Podcasters-\emph{pl}] {
            \includegraphics[width=0.23\textwidth, height=0.15\textwidth]{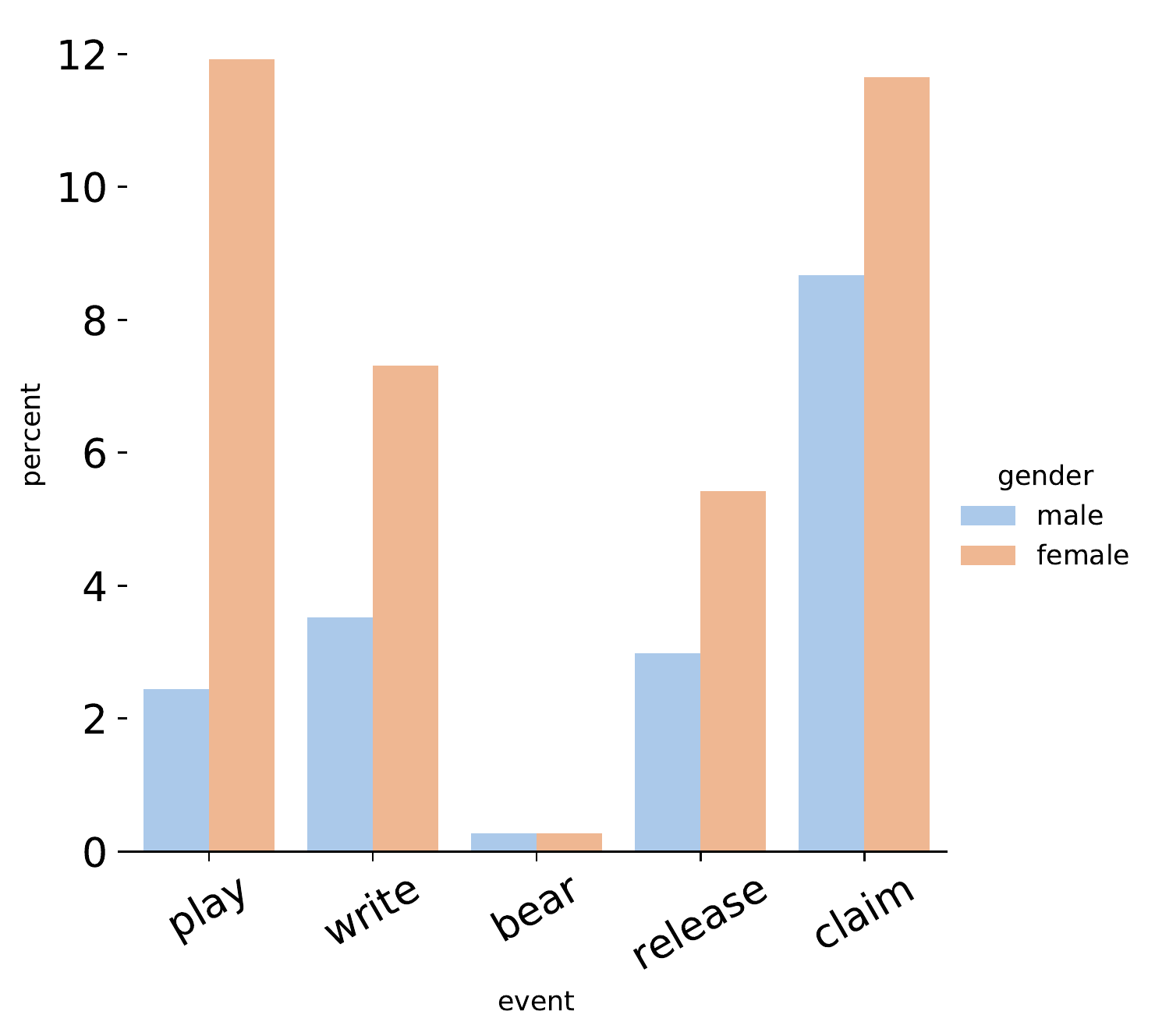}
        }
        \subfloat[Female Podcasters-\emph{pl}] {
                \includegraphics[width=0.23\textwidth, height=0.15\textwidth]{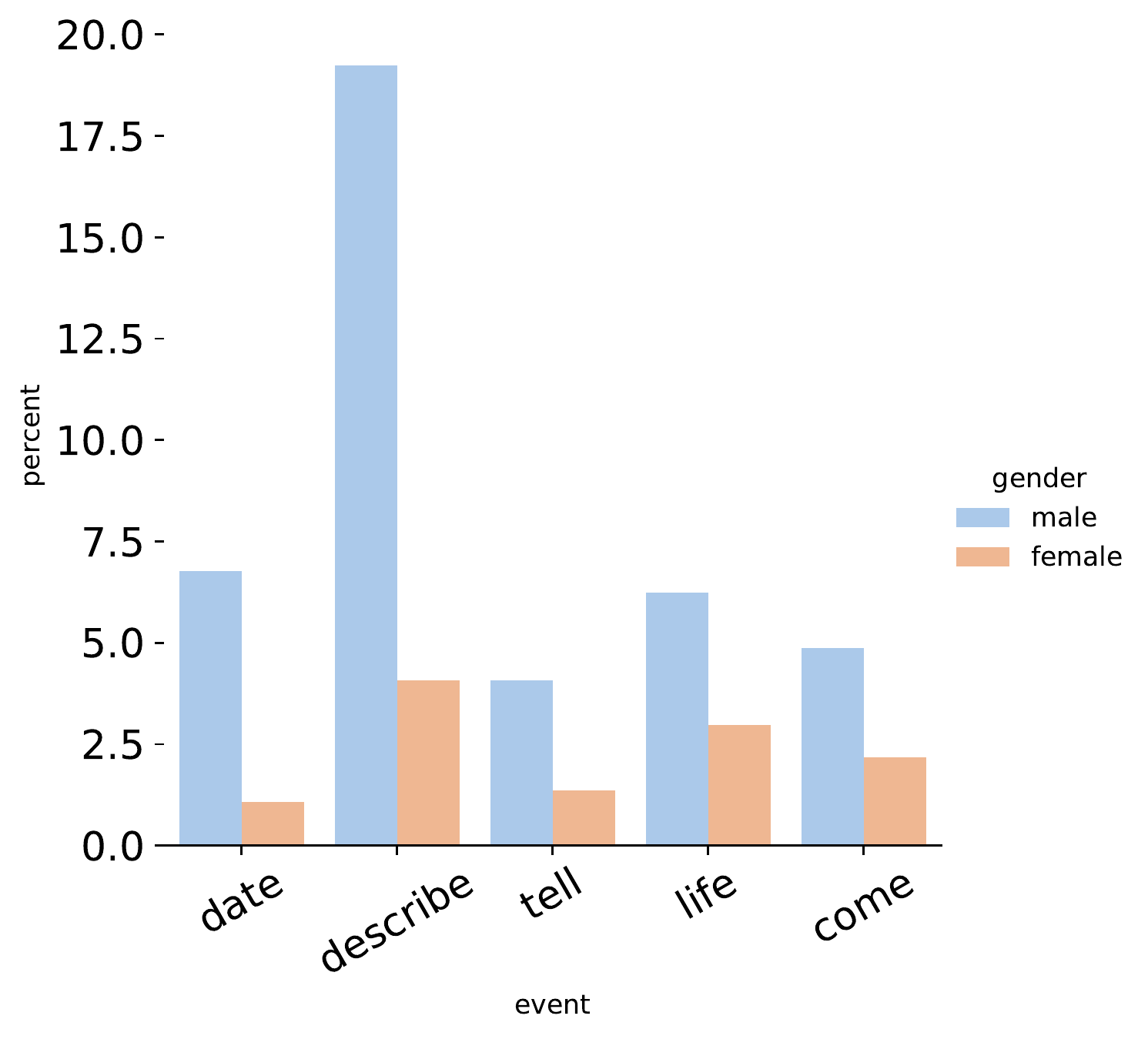}
            }   
        \subfloat[Male Chefs-\emph{pl}]  {
                \includegraphics[width=0.23\textwidth, height=0.15\textwidth]{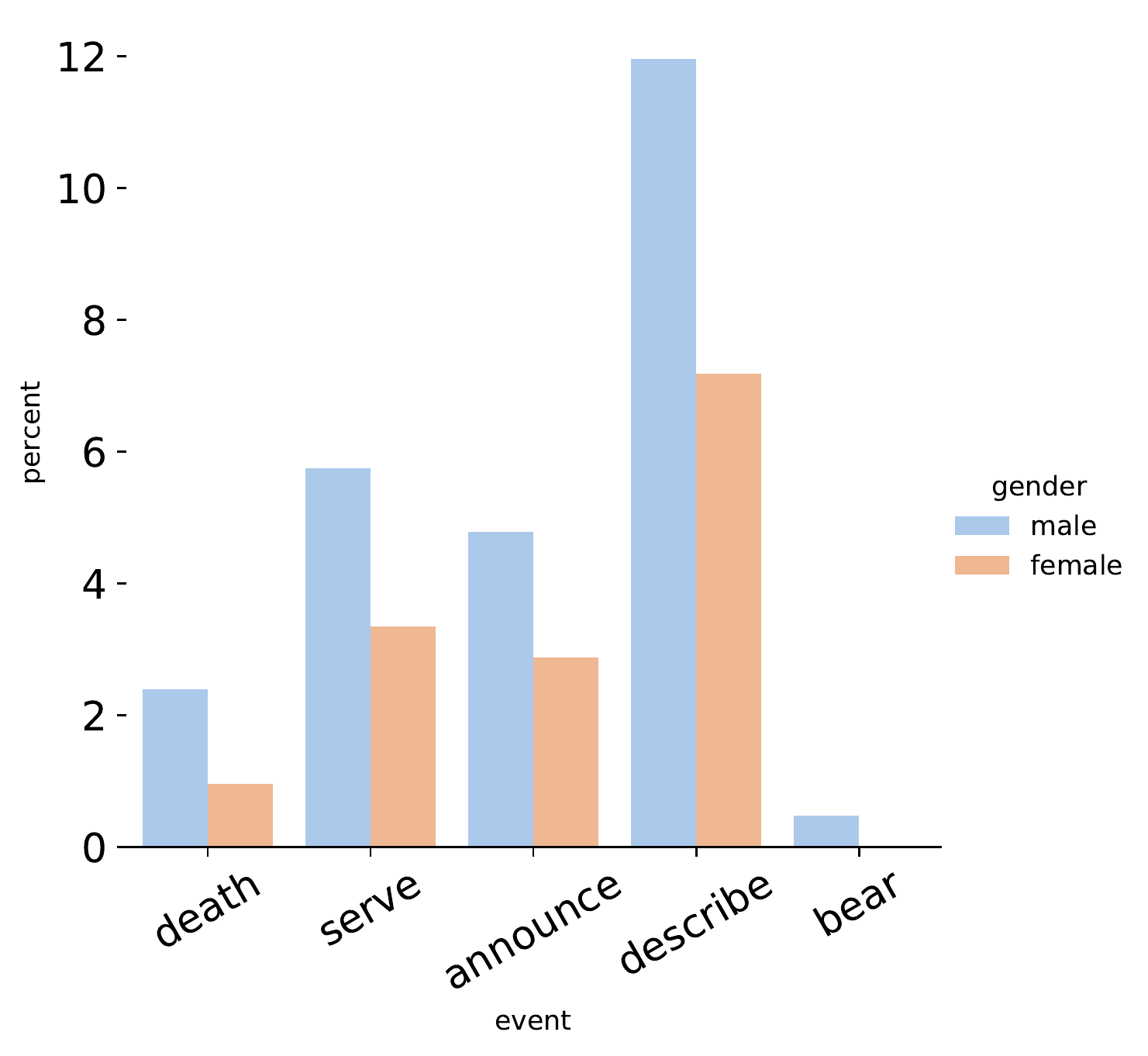}

        }
        \subfloat[Female Chefs-\emph{pl}] {
                \includegraphics[width=0.23\textwidth, height=0.15\textwidth]{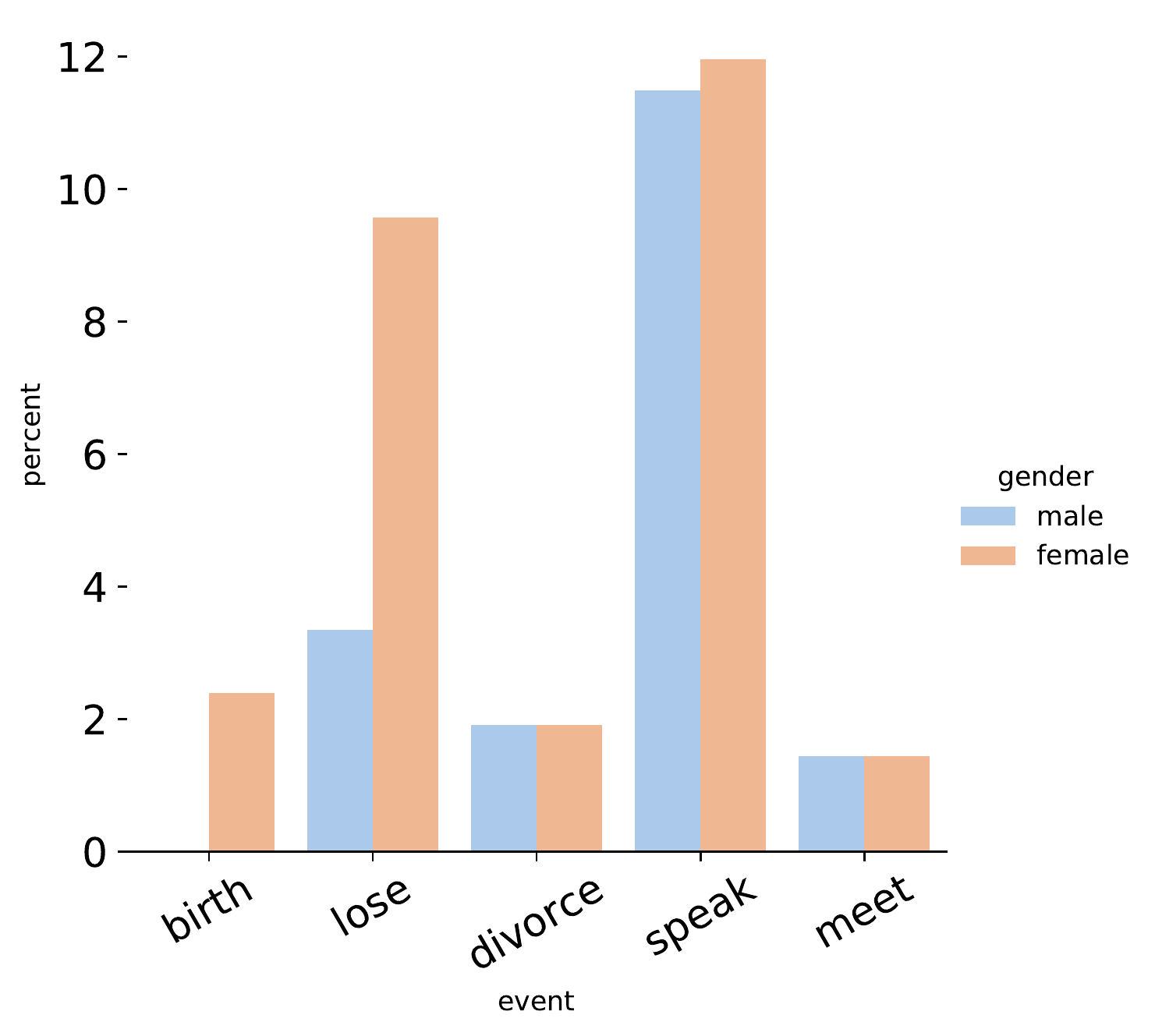}
            }
            
        \subfloat[Male Writers-\emph{pl}] {
            \includegraphics[width=0.23\textwidth, height=0.15\textwidth]{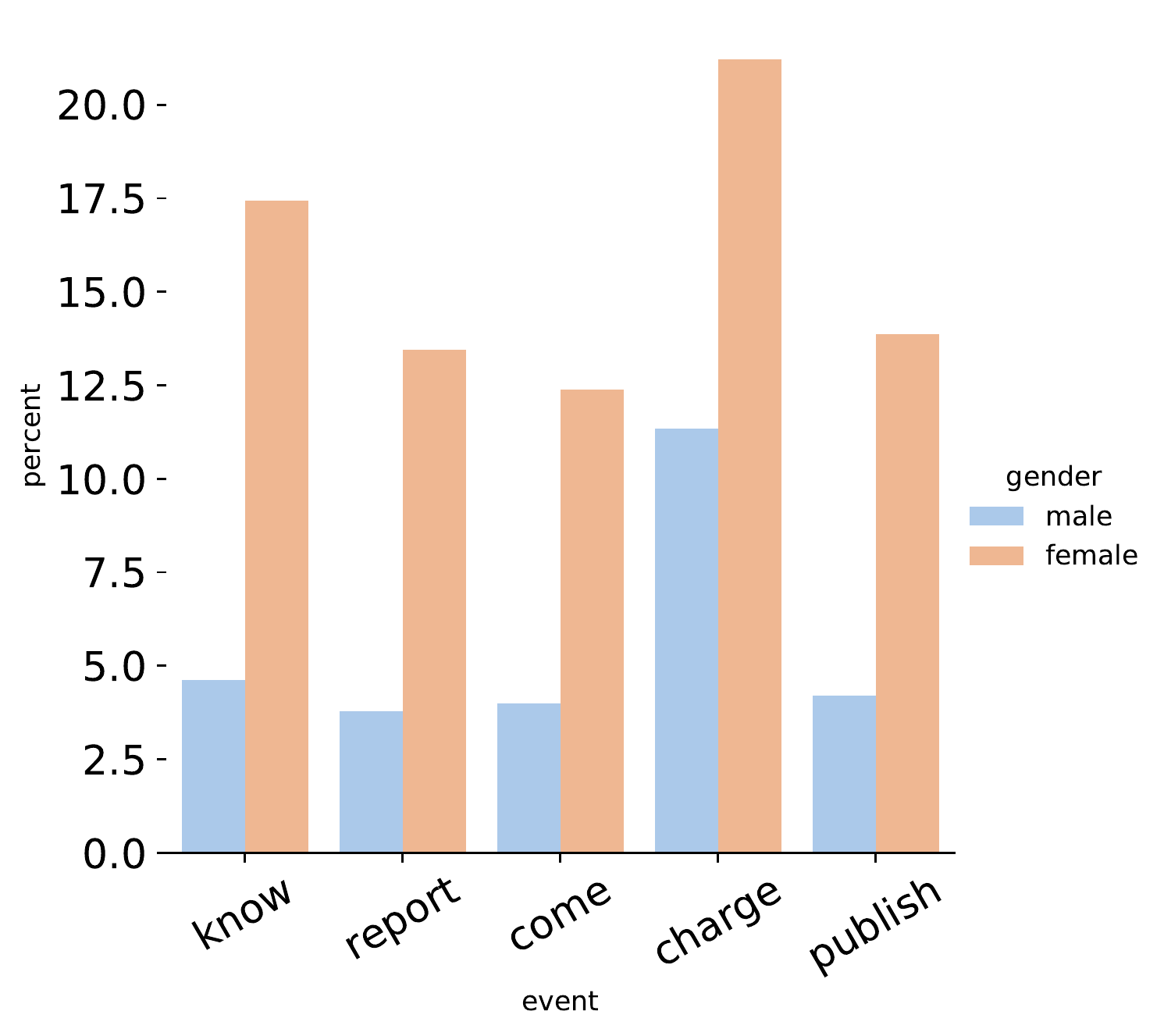}
        }
        \subfloat[Female Writers-\emph{pl}] {
                \includegraphics[width=0.23\textwidth, height=0.15\textwidth]{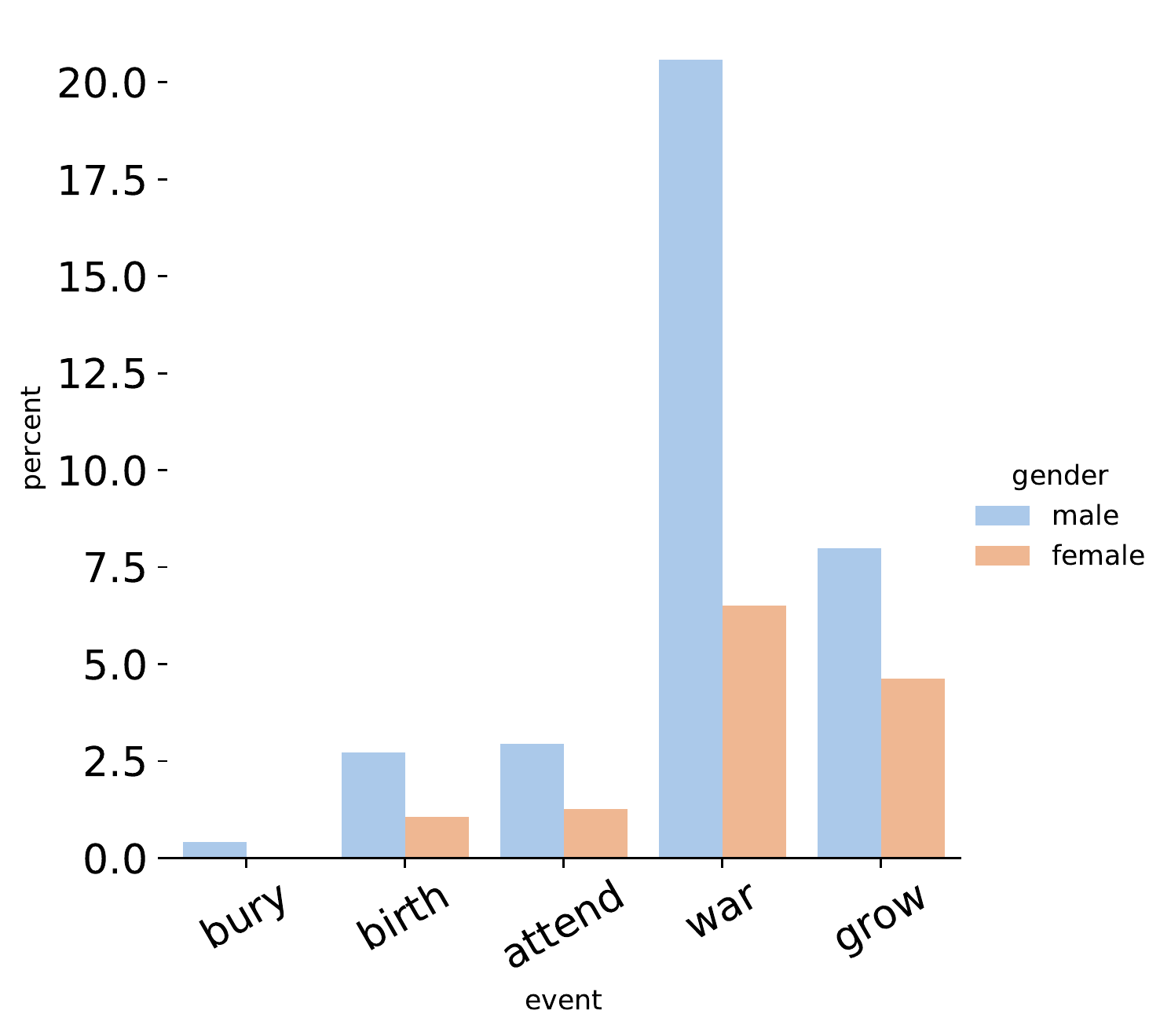}
            }   
        \subfloat[Actors-\emph{pl}]  {
                \includegraphics[width=0.23\textwidth, height=0.15\textwidth]{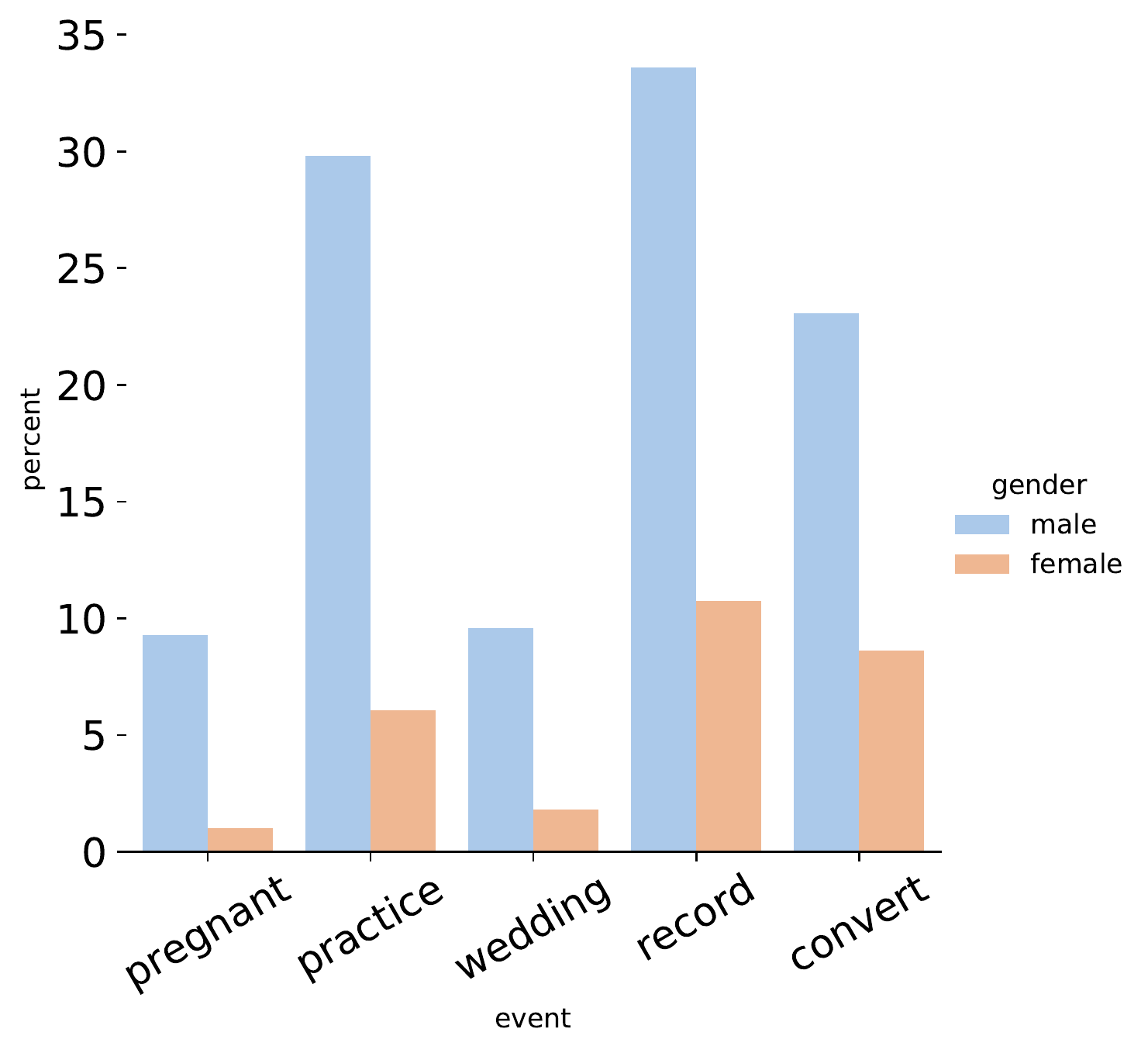}

        }
        \subfloat[Actress-\emph{pl}] {
                \includegraphics[width=0.23\textwidth, height=0.15\textwidth]{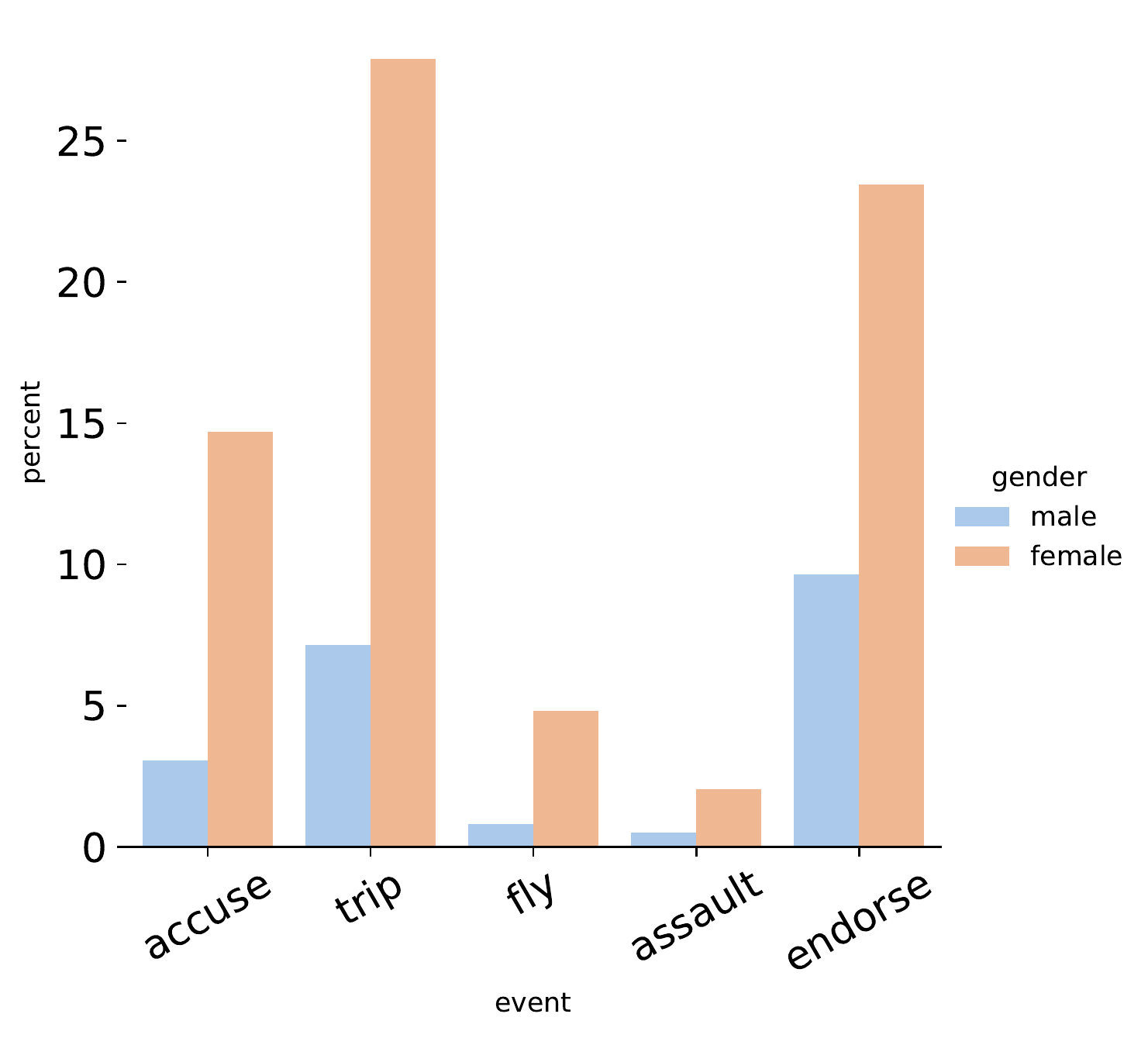}
            }
            }}
    \caption{The percentile of extracted event frequencies. (\emph{c}: \emph{Career} section, \emph{pl}: \emph{Personal Life} section)}
        \label{fig:percent_add}
      
\end{figure*}

We sort all detected events from our corpus by their frequencies in descending order according to Equation 1. Figure~\ref{fig:percent_add} $(a)$-$(l)$ show the percentile of extracted events in the sorted list for another 6 occupations besides the 2 occupations reported in Figure~\ref{fig:percent} for \emph{Career} section. The smaller the percentile is, the more frequent the event appears in the text. These figures indicate that we are not picking events that rarely occur and showcase the significance of our analysis result. Figure~\ref{fig:percent_add} $(m)$-$(x)$ are for \emph{Personal Life} sections across 6 occupations, which show the same trend as for \emph{Career} sections.

\end{document}